\documentclass[aps,pre,preprint,superscriptaddress,longbibliography]{revtex4-1}

\usepackage{algorithm}
\usepackage{algorithmic}
\usepackage{graphicx}
\usepackage{color}
\usepackage{amsmath}
\usepackage{amsthm}
\usepackage{amsfonts}
\begin{document}


\title{Debiased maximum-likelihood estimators of hazard ratios under kernel-based machine-learning adjustment}

\author{Takashi Hayakawa}
\email[]{hayakawa.takashi@nihon-u.ac.jp}
\author{Satoshi Asai}
\affiliation{School of Medicine, Nihon University, Itabashi, Tokyo, 1738610, Japan}


\date{\today}

\begin{abstract}
Previous studies have shown that hazard ratios between treatment groups estimated with the Cox model are uninterpretable because the unspecified baseline hazard of the model fails to identify temporal change in the risk set composition due to treatment assignment and unobserved factors among multiple, contradictory scenarios. To alleviate this problem, especially in studies based on observational data with uncontrolled dynamic treatment and real-time measurement of many covariates, we propose abandoning the baseline hazard and using kernel-based machine learning to explicitly model the change in the risk set with or without latent variables. For this framework, we clarify the context in which hazard ratios can be causally interpreted, and then develop a method based on Neyman orthogonality to compute debiased maximum-likelihood estimators of hazard ratios, proving necessary convergence results. Numerical simulations confirm that the proposed method identifies the true hazard ratios with minimal bias. These results lay the foundation for developing a useful, alternative method for causal inference with uncontrolled, observational data in modern epidemiology.
\end{abstract}

\keywords{survival analysis; hazard ratio; machine learning; causal inference; kernel method}

\maketitle

\section{Introduction}
The use of hazard ratios to measure the causal effect of treatment has recently come under debate. Although it has been standard practice in epidemiological studies to examine hazard ratios using the Cox proportional hazard model and its modified versions \cite{Lin2020, Bartlett2020}, several studies \cite{Hernan2010, Aalen2015, Martinussen2020} have noted that hazard ratios are uninterpretable with regard to causation (see Martinussen (2022) \cite{Martinussen2022} for a review). In particular, Martinussen {\it et al.} (2020) \cite{Martinussen2020} provided concrete examples of data-generating processes in which the Cox model is correctly specified, but estimated hazard ratios are difficult to interpret. The main difficulty is that time courses of different study populations with different treatment effects are described by the same Cox model with the same set of parameter values. Although a few authors have rebutted the uninterpretability of hazard ratios in the Cox model \cite{Prentice2022, Ying2023, Fay2024}, the issue concerning the unidentifiability with multiple, contradictory scenarios as posed by Martinussen {\it et al}. (2020) \cite{Martinussen2020} remains unresolved.

To address this issue, researchers have sought alternative measures of causal treatment effect, such as differences between counterfactual survival functions or restricted mean survival time \cite{Rufibach2019, Kloecker2020, Snapinn2023}. Methods for applying machine learning to estimate these measures have also been developed \cite{Cui2023, Xu2024, Frauen2025}. However, these measures are only applicable to simple, well-controlled settings, such as randomized clinical trials or observational studies of several covariate-adjusted groups with different baseline treatments, despite such studies making few assumptions about data-generating processes in other respects. Modern epidemiology increasingly requires methods for analyzing large quantities of observational data acquired in a more uncontrolled, dynamic  manner, in which many covariates are measured in real-time. Examples of such data are electronic medical or health records and data acquired by electronic devices for health promotion \cite{Leviton2023}. The marginal structural Cox model potentially used for this purpose \cite{Hernan2001} suffers from the uninterpretability problem described above, and thus, development of an alternative method is required.

The present study proposes a strategy to use a hazard model based on time-dependent treatment variables and covariates, but with no unspecified baseline hazard. Previous studies \cite{Hernan2010, Aalen2015, Martinussen2020} attributed the uninterpretability of hazard ratios to {\it ``selection,''} the fact that less frail subjects (described by unobserved factors) are more likely to remain in the risk set at later stages of the study. Although these authors did not explicitly describe the role of the baseline hazard in the description of this selection process, it is the unspecified baseline hazard that allows the Cox model to be correctly specified for selection processes of many different patterns (see Remark 3-1 below). Thus, we abandon the baseline hazard, and instead exploit the descriptive power of machine learning to capture how the risk set changes over time. This approach can now be implemented due to recent advances in incorporating machine learning into rigorous statistical analysis with effect estimation \cite{vanderLaan2018, Chernozhukov2018}. Prior to this development, machine learning could be used for only outcome prediction in epidemiology, because estimation with machine learning was biased. We thus develop an algorithm that debiases maximum-likelihood (ML) estimators of hazard ratios in the model, using the framework of doubly robust, debiased machine learning (DML) based on Neyman orthogonality (see Ref.\cite{Ahrens2025} for an introductory review). Notably, our algorithm applies to models with latent variables, hence enabling explicit modeling of the selection in the risk set caused by unobserved factors.

The article is organized as follows. In section \ref{sec2_1}, we introduce our problem setting for an observational study with uncontrolled, dynamic treatment and real-time measurement of covariates, together with an exponential, machine-learning-based hazard model used for analysis. In section \ref{sec2_2}, we show that the ML estimator of hazard ratios can be interpreted as a measure of causal treatment effect in this setting (Proposition 1), clarifying required (mostly testable) assumptions. In this argument, we show that the unindentifiability with multiple interpretations is alleviated (Remark 3-1). In section \ref{sec2_3}, we construct Neyman-(near)-orthogonal scores used for debiasing the ML estimators of hazard ratios in our model (Propositions 2 and 3). For these scores, additional convergence results required for DML are also provided in Appendix \ref{appendix_score_regularity}. In section \ref{sec2_5}, we develop an algorithm for computing the debiased estimators of hazard ratios after model selection, suitably adapting the procedure proposed in Ref.\cite{Chernozhukov2018} to our setting. Then, in section \ref{sec3}, we apply the developed method to two sets of clinically plausible simulation data. In the first simulation study described in section \ref{sec3_1}, we show that our estimators appropriately identify the true hazard ratios with minimal bias, under the effect of complicated nonlinear confounding among treatment, comorbidities and outcome. In the second simulation study described in section \ref{sec3_2}, we present a case in which an unobserved factor causes selection in the risk set, resulting in less frail subjects in the later phase of the study. Our debiased estimator based on a model with a latent variable appropriately identify the true hazard ratio with minimal bias. On the basis of these results, in section \ref{sec4}, we discuss multiple advantages of the proposed ML approach over the conventional Cox's maximum-partial-likelihood (MPL) approach as well as its limitations. We list mathematical notations in Table \ref{table1}.  

\section{Theories and Methods} \label{sec2}
\subsection{Exponential parametric hazard model combined with machine learning} \label{sec2_1}
We consider observational studies in which the occurrence or absence of an event of interest in subjects randomly sampled from a large population and indexed by $i(\in \mathcal{I})$ is longitudinally observed over time (indexed by $t\in \mathcal{T}\subset \mathbf{R}$) during a noninformatively right-censored period, $0\leq t\leq C_i$. With a set of time-dependent covariates, collectively denoted by $X_{i,t}(=\{ X_{ij,t}\} _{j\in \mathcal{J}})\in \mathcal{X} \subset \mathbf{R} ^d$ $(d\in \mathbf{N})$, we strive to identify the effect of time-dependent treatment described by a collection of binary variables $A_{i,t}(=\{ A_{ik,t}\} _{k\in \mathcal{K}}\in \mathcal{A}\subset \{ 0,1\} ^{|\mathcal{K}|}$). For clarity of presentation, we assume that at most a single treatment variable can take a value of unity, and the remaining variables must be zero. Consider an exponential hazard model whose conditional hazard is given by
\begin{eqnarray}
h(t|A_{i,t}, X_{i,t})\overset{\mathrm{def}}{=}\exp \left (\theta ^{\prime }A_{i,t}+f(X_{i,t})\right ), \label{exponential_model}
\end{eqnarray}
where $\theta =\{ \theta _k\} _{k\in \mathcal{K}}\in \Theta \subset \mathbf{R}^{|\mathcal{K}|}$ is a set of parameters corresponding to the natural logarithm of the hazard ratio of the untreated samples to samples treated to different extents, and the function $f$ describes the risk variation due to the given set of covariates, which we adjust using a machine learning model (i.e., a set of functions $\mathcal{M}$ in which $f$ lies). In addition, hereinafter, $(\cdot)^{\prime }$ denotes the transposition of vectors and matrices. Note that the above model lacks the baseline hazard as a function of $t$ (the time elapsed after enrollment). The covariate $X_{i,t}$ can include temporal information, such as date and durations of suffered disorders, but not $t$ itself.

Suppose that a subset of subjects (indexed by $i$) experiences an event of interest at time $T_i$ in the observation period, and the other subjects experience no such event. The full log-likelihood for this observation is then given by  
\begin{eqnarray}
\ln Q_h(T|A, X)=\sum _{i\in \mathcal{I}}\left \{ I_{[0,C_i]}(T_i)(\theta ^{\prime }A_{i,T_i}+f(X_{i,T_i}))-\int _{0}^{T_i\land C_i} \exp\left (\theta ^{\prime }A_{i,t}+f(X_{i,t})\right ) dt\right \} , \label{log_likelihood}
\end{eqnarray}
(see, e.g., Ref.\cite{Ren2011}), where $I_{[0, C_i]}(T_i)$ is the indicator function that returns unity for $T_i\in [0, C_i]$ and zero otherwise. To see the meaning of Eq.(\ref{log_likelihood}), we consider how the hazard model describes the occurrence of an event in a small interval $(t, t+\delta t]$. Assuming for the moment that (i) $X_{i,s}$ is continuous with respect to $s$ and (ii) $A_{i,s}=const$ over $s\in [t, t+\delta t]$, the likelihoods of occurrence and non-occurrence of an event in this period conditioned on event-free survival at time $t$ are $h(t|A_{i,t}, X_{i,t})\delta t$ and $1-h(t|A_{i,t},X_{i,t})\delta t$, respectively, up to the first order in $\delta t$. Simple computations show that the log-likelihood for this binary observation is 
\begin{eqnarray}
\ln Q_h(I_{(t,t+\delta t]}(T_i)|T_i>t, A_{i,t},X_{i,t})&=&I_{(t,t+\delta t]}(T_i)\ln h(t|A_{i,t}, X_{i,t})\delta t \nonumber \\
 &&\hspace{-2.0cm}+(1-I_{(t,t+\delta t]}(T_i))\ln (1-h(t|A_{i,t}, X_{i,t})\delta t)+ O(\delta t^2), \nonumber \\ 
 &&\hspace{-1.0cm}=I_{(t,t+\delta t]}(T_i) (\theta ^{\prime }A_{i,t}+f(X_{i,t}))-\exp \left (\theta ^{\prime }A_{i,t}+f(X_{i,t})\right )\delta t \nonumber \\
 &&\hspace{-1.5cm}+O(\delta t^2)+const. \label{infinitesimal_hazard_likelihood}
\end{eqnarray} 
The time integration of Eq.(\ref{infinitesimal_hazard_likelihood}) results in the summand of Eq.(\ref{log_likelihood}) up to a constant independent of $\theta $ and $f$.

We assume that the two assumptions described above hold almost surely in the $\delta t\rightarrow +0$ limit, because $\{A_{it}\} _t$ and $\{ X_{it}\} _t$ are continuous with respect to time except for finite numbers of time points in real applications. Here, however, we make it clear in advance that we do not go into the details about the necessary conditions on the data-generating stochastic process, $\{ A_{it}, X_{it}\} _t$. We do not specify the filtered probability space for the process, but rather assume the existence of the marginal (conditional) probability measures on the measurable spaces $(\mathcal{A}\times \mathcal{X}\times \mathcal{T}, \mathfrak{M} _\mathcal{A} \times \mathfrak{M} _{\mathcal{X}} \times \mathfrak{M}_\mathcal{T} )$, and the integrability of quantities of interest with respect to these measures. For the model $\mathcal{M}$ on the base space $(\mathcal{X} ,\mathfrak{M} _{\mathcal{X}})$ for covariates, we consider only reproducing-kernel Hilbert spaces (RKHS) $\mathcal{H} _k$ associated with a positive-semidefinite bounded kernel $k(\cdot, \cdot)$ \cite{Berlinet2011}. We refer readers to Ref.\cite{Fukumizu2013} which provides a basic tool for our theoretical analysis of kernel-based machine learning.

\subsection{Causal inference and ML estimation} \label{sec2_2}
This section clarifies the causal interpretation of the estimation results in the model described in the previous section and discusses the necessary assumptions that support this interpretation. First, we make the conventional assumptions of consistency, no unmeasured confounder (ignorability), and positivity.

\noindent {\bf Assumption 1 (consistency).} {\it For any counterfactual treatment schedule $a$, we have, $X^a_{(-\infty, t]}=X_{(-\infty,t]}|A_{(-\infty ,t]}=a_{(-\infty,t]}$ and $N_{(-\infty ,t]}^a=N_{(-\infty ,t]}|A_{(-\infty ,t]}=a_{(-\infty,t]}$, for counting processes $N_{t}=I_{(-\infty , t]}(T)$ and $N_{t}^a=I_{(-\infty , t]}(T^a)$.}

\noindent {\bf Assumption 2 (no unmeasured confounder).} {\it For some $\epsilon >0$, we have, for any $t$ and any counterfactual treatment schedule $a$, }
\begin{eqnarray} 
T^a \mathop{\perp\!\!\!\!\perp}  A_{[t, t+\epsilon ]}&|&T^a>t, X_{(-\infty, t+\epsilon ]}, A_{(-\infty  , t)}.
\end{eqnarray}
\noindent {\bf Assumption 3 (positivity) [optional].} {\it For any $t>0$ and any $a\in \mathcal{A}$, we have $P(A_{t}=a|T>t, X_t)>0$.}
 
Here, the set subscript, such as $X_{(-\infty , t]}$, denotes the collection of all variables with subscripts included in the set. Throughout the paper, $P$ denotes the marginal (conditional) probability measures of the argument variables derived from the data-generating process. We have used variables for $-\infty <t<0$ for a technical reason that $A_0$ and $X_0$ must be determined on the basis of the past information. This may be replaced by some finite interval before the enrollment at $t=0$ for which the information about treatment and covariate is available.  

Next, in addition to the above, we make the following assumption about the regularity of hazard. 

\noindent {\bf Assumption 4 (regularity of hazard).} 
\begin{eqnarray}
&&\lim _{\delta t\rightarrow +0}\frac{1}{\delta t}P(I_{(t, t+\delta t]}(T)=1|T>t, A_{(-\infty , t+\delta t]}, X_{(-\infty, t+\delta t]})\nonumber \\
&=&\lim _{\delta t\rightarrow +0}\frac{1}{\delta t}P(I_{(t, t+\delta t]}(T)=1|T>t, A_{(-\infty ,t]}, X_{(-\infty ,t]}) \nonumber \\
 &<&\infty .
\end{eqnarray}
{\it holds almost surely. }

\noindent{\bf Remark 1-1.} Assumption 1 is a natural assumption without which we have two different processes for the same treatment schedule.

\noindent {\bf Remark 1-2.} Assumption 2 is stating that the assignment of treatment at each time point $t$ can be considered as randomized one conditioned on the value of covariate. In the analysis of electronic medical records, for example, if the medical practitioners decide their treatment on the basis of evidence recorded as covariates and those covariates are measured in real time, this assumption is valid. It should be noted that if the decision is based on past records, a past variable should be combined with the present variable to form a vectorized covariate. Causal inference with the same assumption with discretized timesteps has been performed with marginal structural Cox models (see Ref.\cite{Hernan2001,Yang2014} for examples). 

\noindent {\bf Remark 1-3.} Assumption 3 is a technical assumption required for debiasing with inverse-treatment-probability weights (see Proposition 3 and Remark 5-1 for our model and Ref.\cite{Robins2000} for marginal structural Cox models). If medical practioners never choose a particular treatment option for a value of covariate, this assumption is violated. Thus, if one analyzes the effect of a treatment considered as a taboo for comorbidities included in the set of covariates (for example, anticholinergics for angle-closure glaucoma), they need to take an extra care of this assumption. This assumption can be omitted if one uses Proposition 2 for estimation in our study. 

\noindent {\bf Remark 1-4.} Assumption 2 essentially implies that the treatment at time $t$ does not affect the time evolution of covariate $X$ for a subsequent period $(t, t+\epsilon ]$ as seen through the following argument. First, we have 
 \begin{eqnarray}
&&P(I_{(t, t+\epsilon ]}(T^a)|T^a>t, X_{(-\infty , t+\epsilon]}, A_{(-\infty ,t)}=a_{(-\infty , t)})\nonumber \\
&=&P(I_{(t, t+\epsilon]}(T^a)|T^a>t, X_{(-\infty , t+\epsilon]}, A_{(-\infty , t+\epsilon]}=a_{(-\infty , t+\epsilon ]})\nonumber \\
&=&P(I_{(t, t+\epsilon]}(T)|T>t, X_{(-\infty , t+\epsilon]}, A_{(-\infty, t+\epsilon]}=a_{(-\infty , t+\epsilon ]}),  \label{ignorability_conseq}
\end{eqnarray}
where the first and second equalities are due to no unmeasured confounder and consistency, respectively. If the time evolution of covariates is immediately affected by the treatment schedule, and if outcome is immediately affected by the covariates, then under the conditioning on $X_{[t, t+\epsilon]}$, $X^a_{[t, t+\epsilon]}$ depends on the value of $A$, and hence, $T^a$ depends on $A$, which contradicts the assumption. The removal of such instantaneous interactions among treatment, covariates and outcome also makes Assumption 4 reasonable.

In addition to Assumptions 1--4, we assume that the model in Eq.(\ref{exponential_model}) is correctly specified. More precisely, we make the following four assumptions:\\
\noindent {\bf Assumption 5 (hazard conditionally independent of past treatment and covariate).} {\it For any $t>0$, $a\in \mathcal{A}$, $x\in \mathcal{X}$ and $E_A, E_X\subset (-\infty, t)$, we almost surely have},
\begin{eqnarray}
\lim _{\delta t\rightarrow +0}\frac{P(I_{(t,t+\delta t]}(T)=1|T>t, A_t=a, X_t=x)}{P(I_{(t,t+\delta t]}(T)=1|T>t, A_t=a, X_t=x, A_{E_A}, X_{E_X})}=1. 
\end{eqnarray}

\noindent {\bf Assumption 6 (homogeneous treatment effect).} {\it For any $a, a^{\prime}\in \mathcal{A}$ and any $x\in \mathcal{X}$, the hazard contrast,
\begin{eqnarray}
\lim _{\delta t\rightarrow +0}\frac{P(I_{(t,t+\delta t]}(T)=1|T>t, A_t=a, X_t=x)}{P(I_{(t,t+\delta t]}(T)=1|T>t, A_t=a^{\prime }, X_t=x)},
\end{eqnarray}
takes a constant value regardless of the value of $x$.}

\noindent {\bf Assumption 7 (time-homogeneous hazard):} {\it The hazard is independent of its timing during the observation period. Mathematically, for any $t,t^{\prime }\geq 0$, and any fixed $a\in \mathcal{A}$ and $x\in \mathcal{X}$,}
\begin{eqnarray}
\lim _{\delta t\rightarrow +0}\frac{P(I_{(t,t+\delta t]}(T)=1|T>t, A_t=a, X_t=x)}{P(I_{(t^{\prime },t^{\prime }+\delta t]}(T)=1|T>t^{\prime }, A_{t^{\prime}}=a, X_{t^{\prime }}=x)} =1.
\end{eqnarray}
\noindent {\bf Assumption 8 (correctly specified machine-learning model):} {\it There exist unique $\theta ^*\in \mathbf{R}$ and $f^*\in \mathcal{M}$ that satisfy
\begin{eqnarray}
\mathrm{E} \left [\ln Q_{h_{\theta ^*, f^*}}(T|A,X)\right ]=\sup _{\theta \in \mathbf{R}, f\in \mathfrak{B}_\mathcal{X}} \mathrm{E} \left [\ln Q_{h_{\theta ,f}}(T|A,X)\right ], 
\end{eqnarray}
where $\mathfrak{B} _\mathcal{X}$ is the set of Borel-measurable functions.}

 \noindent {\bf Remark 2.} Assumptions 5--7 can be validated by extending the model to incorporate the effect of past treatment and covariates, inhomogeneous treatment effect, and time inhomogeneity, respectively. The original and extended models can then be compared in terms of, for example, Bayesian model evidence (BME) (see, e.g., Chapter 3 of Ref.\cite{Bishop2006}). If Assumption 5 is violated, the definition of the treatment variables and covariates may be extended so that the current value retains past information. This is the same strategy as employed for designing variables in a marginal structural Cox model \cite{Hernan2001}. If Assumption 6 is violated, the model may be extended by incorporating covariate-dependent heterogeneous treatment effect. Violation of Assumption 7 indicates that variations in covariate values do not account for the temporal change in the risk set. In this case, additional covariates or extending the model with latent variables (as described below) may be considered. Finally, Assumption 8 is an {\it a priori} assumption that stems from the fact in statistical learning that inference in the function space requires a regularity assumption.

\noindent {\bf Proposition 1.} {\it Let the maximizer of the expected likelihood be
\begin{eqnarray}
\theta ^*, f^*=\mathrm{arg} \max _{\theta \in \mathbf{R}, f\in \mathcal{M}}\mathrm{E} [\ln Q_h(T|A, X)].
\end{eqnarray}
Under Assumptions 1--8, suppose two possible treatment schedules $a, a^{\prime }$ such that $a_{(-\infty , t)}=a^{\prime }_{(-\infty , t)}$ holds, and that $a_{k,s}=1$ and $a_{s}^{\prime }=0$ (or $a_{k^{\prime },s}^{\prime }=1$) holds for $s\in [t, t+\epsilon ]$ for some $\epsilon >0$. Then, we have, for any sample path $x_{(-\infty , t+\delta t]}$ for covariate, }
\begin{eqnarray}
&&\hspace{-0.5cm}\theta _k^*(-\theta _{k^{\prime }}^*)=\nonumber \\
&&\hspace{-0.5cm}\lim _{\delta t\rightarrow +0}\ln \frac{ P(I_{(t,t+\delta t]}(T^a)=1|T^a, T^{a^{\prime }}>t, A_{(-\infty , t)}=a_{(-\infty , t)}, X_{(-\infty , t+\delta t]}=x_{(-\infty , t+\delta t]})}{P(I_{(t,t+\delta t]}(T^{a^{\prime }})=1|T^a,T^{a^{\prime }}>t, A_{(-\infty , t)}=a^{\prime }_{(-\infty , t)}, X_{(-\infty , t+\delta t]}=x_{(-\infty , t+\delta t]})} . \label{inst_causal_effect}
\end{eqnarray}

\begin{proof}
see Appendix \ref{appendix_prop1}.
\end{proof}

\noindent {\bf Remark 3-1.}  Suppose that the entire covariate $X_t$ is unobserved. In this case, the modified version of the above proof indicates that a hazard model expanded with respect to time:
\begin{eqnarray}
h(t|A_t)\overset{\mathrm{def}}{=}\exp (\theta (t)^{\prime }A_t+f(t))
\end{eqnarray}
admits the optimal solution 
\begin{eqnarray}
\theta ^*(t)&=&\ln \lim _{\delta t\rightarrow +0}\frac{\int _{\mathcal{X}}P(I_{(t,t+\delta t]}(T)=1|T>t, A_{k,t}=1, X_t)dP(X_t|A_{k,t}=1, T>t)}{\int _{\mathcal{X}}P(I_{(t,t+\delta t]}(T)=1|T>t, A_{t}=0, X_t)dP(X_t|A_{t}=0, T>t)}, \nonumber \\
f^*(t)&=&\ln \lim _{\delta t\rightarrow +0}\frac{1}{\delta t}\int _{\mathcal{X}}P(I_{(t,t+\delta t]}(T)=1|T>t, A_t=0, X_t)dP(X_t|A_{t}=0, T>t), \nonumber 
\end{eqnarray}
where one can interpret $\exp (f^*(t))$ as a baseline hazard. The example of the uninterpretability of hazard ratios presented by Martinussen et al. \cite{Martinussen2020} was for this setting. They assumed time-independent $A_t$ and $X_t$. Their point was that multiple combinations of $dP(X_t|A_t, T>t)$ and $\lim _{\delta t\rightarrow +0}P(I_{(t,t+\delta t]}(T)=1|T>t, A_t, X_t)/\delta t$ yields the same solution for $\theta (t)$.

This ambiguity has been removed in our setting. We assume that $\lim _{\delta t\rightarrow +0}P(I_{(t,t+\delta t]}$ $(T)=1|T>t, A_{t}, X_t)/\delta t$ is independent of specific time points of the observation period, which is biologically natural. Then, time-dependence of solutions for $\theta $ and $f$ indicates lack of observation of relevant factors and encourages us to seek for a better model. This proof mechanism does not work in the Cox model, because the Cox model does not estimate the baseline hazard. Although our framework cannot exclude a scenario in which temporal changes in the distribution of unobserved factors miraculously keep $\theta ^*(t)$ and $f^*(t)$ constant over time, this is not a generic case.

\noindent {\bf Remark 3-2.} The quantity in Eq.(\ref{inst_causal_effect}) is a covariate-adjusted hazard contrast for two counterfactual treatments that branch at time $t$, so it can be interpreted as a measure of causal effect. In the setting discussed in Ref.\cite{Hernan2001}, this corresponds to the treatment effect measured in the next month (for which covariates at each month can be regarded as baseline covariates \cite{vanderLaan2005}). 

\noindent {\bf Remark 3-3.} See Remarks 8-1 and 8-2 in Appendix \ref{appendix_causal_interpretation} for more consideration on causal interpretation of hazard ratios.

\subsection{Debiasing ML estimators} \label{sec2_3}
Next, we assume that we have consistent estimators for $\theta$ and $f$ that are the maximizers of the empirical log-likelihood in Eq.(\ref{log_likelihood}) in the presence of a suitable regularizer. Although a large body of the machine-learning literature presents such consistent estimators, they are biased in the sense that we cannot expect $\sqrt{n} \|\widehat{\theta }-\theta ^*\| _2, \sqrt{n} \| \widehat{f}-f^*\| _{\mathcal{M}}\overset{p}{\rightarrow }0$ for the number of subjects $n(\rightarrow \infty)$. This situation prohibits making a decision on the significance of estimated results. However, we can apply to this problem debiased machine learning based on Neyman orthogonality and developed by Chernozhukov {\it et al}. (2018) \cite{Chernozhukov2018}. Chernozhukov {\it et al}. (2018) describe how to systematically debias ML estimators and other M-estimators. Applying their idea to our problem setting yields debiased estimators of $\theta ^*$.

\noindent {\bf Definition 1 (Neyman near-orthogonal scores).} {\it Suppose that an estimator $\widehat{\theta } (\in \Theta \subset \mathbf{R} ^d)$ of quantities of interest $\theta ^*$ is given as a zero of the empirical average of score functions $\phi: \mathcal{D} \times \Theta \times \mathcal{N} \rightarrow \mathbf{R} ^d$ for i.i.d. data $D_i\in \mathcal{D}$: 
\begin{eqnarray}
\widehat{E}[\phi (D; \widehat{\theta }, \widehat{\eta } )]\overset{\mathrm{def}}{=}\frac{1}{n} \sum _{i} \phi (D_i; \widehat{\theta }, \widehat{\eta } )=0,
\end{eqnarray}
where $\widehat{E}$ denotes empirical average and the third argument of $\phi $ is a (possibly infinite-dimensional) nuisance parameter and an estimator $\widehat{\eta }$ of an optimal value $\eta ^* \in \mathcal{N}$ is used. Also suppose that there exist convex neighbourhoods of $\eta ^*$, $\mathcal{N} _n\subset \mathcal{N}$ to which $\widehat{\eta }$ belongs with high probabilities converging to one. Then, $\phi $ is said to be ``Neyman near-orthogonal,'' if it satisfies
\begin{eqnarray}
\mathrm{E} [\phi (D; \theta ^*, \eta ^*)]&=&0,\\
\partial _{\eta } \mathrm{E} [\phi (D; \theta ^*, \eta ^*)][\eta -\eta ^*]&\leq &\epsilon _n\sim o(n^{-1/2})\ (^{\forall }\eta \in \mathcal{N}_n),
\end{eqnarray}
where, in the second line, $\partial _{\eta }(\cdot )[\eta -\eta ^*]$ denotes the Gateaux derivative operator (see Table \ref{table1} and Ref.}\cite{Hille1974}. {\it for definition). If the above condition holds for $\epsilon _n=0$, $\phi $ is said to be ``Neyman orthogonal''.}

Given a few conditions on the regularity of score functions and the convergence of nuisance parameters, Chernozhukov {\it et al}. \cite{Chernozhukov2018} proved $\sqrt{n} \|\widehat{\theta }-\theta ^*\|\overset{p}{\rightarrow }0$. In the following, we construct such orthogonal scores from the log-likelihood in Eq.(\ref{log_likelihood}). For this purpose, we modify the following lemma proven by Chernozhukov {\it et al}. \cite{Chernozhukov2018}.

\noindent {\bf Lemma 1 (Lemma 2.5 of Chernozhukov et al. 2018: Neyman orthogonal scores derived from log-likelihood).} {\it Consider a ML problem, 
\begin{eqnarray}
\theta ^*, f^*=\arg \min _{\theta \in \Theta , f\in \mathcal{M}} \mathrm{E} [\ell (D; \theta , f)], 
\end{eqnarray}
where $\ell $ is a sample-wise negative log-likelihood function and $\mathcal{M}$ is a convex set of high-dimensional vectors. Define $f_{\theta }$ with 
\begin{eqnarray}
f_{\theta }=\arg \min _{f\in \mathcal{M}} \mathrm{E} [\ell (D; \theta , f)], 
\end{eqnarray}
and a convex set $\mathcal{N}$ of mappings of $\Theta $ into $\mathcal{M}$ with an optimal solution in this set $\eta ^*(\theta )=f_{\theta }$. Further suppose that for each $\eta \in \mathcal{N}$, the function $\theta \mapsto \ell (D; \theta, \eta (\theta ))$ is continuously differentiable almost surely. Then, under mild regularity conditions, the score $\phi $: 
\begin{eqnarray}
\phi (D; \theta , \eta )=\frac{d\ell (D; \theta , \eta (\theta ))}{d\theta }
\end{eqnarray} 
is Neyman orthogonal at $(\theta ^*, \eta ^*)$. Here, the differentiation on the right-hand side of the above is the full derivative.
}

For a finite-dimensional $f$, the above lemma with the application of the implicit function theorem to $\partial _{\theta } E[\ell (D; \theta , f)]=0$ (or its empirical version) yields the following orthogonal score (section 2.2.1 of Ref.\cite{Chernozhukov2018}): 
\begin{eqnarray}
\phi (D; \theta , (f,\mu))=\partial _\theta \ell (D; \theta ,f)-\mu \partial _f\ell (D; \theta , f),
\end{eqnarray}
with $\mu ^*=E[\partial _\theta \partial _f\ell |_{\theta ^*, f^*}]E[\partial _f\partial _f\ell |_{\theta ^*, f^*}]^{-1}$ and its estimator $\widehat{\mu }=\widehat{E}[\partial _\theta \partial _f\ell |_{\widehat{\theta }, \widehat{f}}]\widehat{E}[\partial _f\partial _f\ell |_{\widehat{\theta }, \widehat{f}}]^{-1}$ as long as the Hessians with respect to $f$ are invertible.

In the following, for our high-dimensional machine-learning applications, we develop an analogous approach with an extra care.  
 
\noindent {\bf Proposition 2 (Neyman near-orthogonal scores for ML in the present model).} {\it Suppose that the model $\mathcal{M}$ is given as a reproducing-kernel Hilbert space $\mathcal{H} _k$ associated with a bounded positive-semidefinite kernel $k$ on $\mathcal{X}\times \mathcal{X}$. Then, for the sample-wise negative log-likelihood $\ell $ of the present study, the gradients $\partial _f\ell $, $\partial _f\partial _{\theta _k}\ell $ and the Hessian $\partial _f\partial _f\ell $ can be identified with elements of $\mathcal{H}_k$ and a linear operator from $\mathcal{H}_k$ into $\mathcal{H}_k$ in the following sense: 
\begin{eqnarray}
\ell (D; \theta , f+h)&=&\ell (D; \theta , f)+(\partial _f\ell |_{\theta ,f}, h)_{\mathcal{H}}+O(\|h\| _{\mathcal{H}_k}^2), \nonumber \\
\partial _{\theta _k}\ell (D; \theta , f+h)&=&\partial _{\theta _k}\ell (D; \theta , f)+(\partial _f\partial _{\theta _k}\ell |_{\theta ,f}, h)_{\mathcal{H}}+O(\|h\| _{\mathcal{H}_k}^2), \nonumber \\
\partial _f\ell |_{\theta , f+h}&=&\partial _f\ell |_{\theta ,f}+\partial _f\partial _f\ell |_{\theta ,f} h +O(\| h\| _{\mathcal{H}_k}^2).
\end{eqnarray}
If Assumption 9 (stated below) holds, for $\zeta _n=cn^{-\alpha }$ with constants $c>0$ and $\alpha >\frac{1}{2}-\beta $, the score function, 
\begin{eqnarray}
\phi (D; \theta , (f, H))=\partial _\theta \ell (D; \theta ,f)-H_{\theta f}(H_{ff}+\zeta _n)^{-1}\partial _f\ell (D; \theta , f), 
\end{eqnarray}
is Neyman near-orthogonal for the shrinking set of nuisance parameters $\mathcal{N} _n=\mathcal{B} _n\times \mathcal{S} _n$ (defined below) with the true value $H^*=\mathrm{E} [\partial _{(\theta, f)} \partial _{(\theta ,f)}\ell |_{\theta ^*, f^*}]$ and empirical estimators $\widehat{H} =\widehat{\mathrm{E}} [\partial _{(\theta ,f)}\partial _{(\theta ,f)}\ell |_{\widehat{\theta }, \widehat{f}}]$. Here, $H_{\theta f}$ and $H_{ff}$ are blocks of the positive-semidefinite compact linear operator $H: \mathbf{R} ^{|\mathcal{K}|}\times \mathcal{H} _k \rightarrow \mathbf{R} ^{|\mathcal{K}|}\times \mathcal{H} _k$ with  
\begin{eqnarray}
\left (\begin{array}{c} H_{\theta f} \\ H_{ff} \end{array}\right )\ :\  \psi \in \mathcal{H} _k\  \mapsto \ H \left (\begin{array}{c} 0 \\ \psi \end{array}\right ).
\end{eqnarray}
The other blocks are similarly defined.
}  

\begin{proof}
see Appendix \ref{appendix_prop2}.
\end{proof}

\noindent {\bf Assumption 9} {\it Estimator $(\widehat{\theta }, \widehat{f})$ falls into $\mathcal{B} _n\overset{\mathrm{def}}{=}\{ (\theta , f)\ |\ \| \theta -\theta ^*\| _2, \| f-f^*\| _{\mathcal{H} _k} \leq c_1n^{-\beta }\} $ for some positive constants $c_1$ and $\beta (<1/2)$, with probabilities converging to one for $n\rightarrow \infty $. Estimator $\widehat{H}=\widehat{E}[\partial _{(\theta ,f)}\partial _{(\theta ,f)}\ell |_{\widehat{\theta },\widehat{f}}]$ falls in $\mathcal{S} _n\overset{\mathrm{def}}{=}\{ H\ |\ $} positive semidefinite compact op. from $\mathbf{R} ^{|\mathcal{K}|}\times \mathcal{H}_k$ into $\mathbf{R} ^{|\mathcal{K}|}\times \mathcal{H}_k$ {\it s.t. $\| H-H^*\| \leq c_2n^{-\beta }\} $ with probabilities converging to one for $n\rightarrow \infty $. It is further assumed that, for each $k\in \mathcal{K}$, $H_{f\theta _k}^*=H_{ff}^*\rho _k$ holds for some element $\rho _k$ of $\mathcal{H}_k$.}

\noindent {\bf Remark 4-1.} It can be argued that Assumption 9 is likely to hold (see Appendix \ref{appendix_assumption9}). However, we leave this as a conjecture, because its proof needs an explicit argument about the details of the data-generating stochastic process. 

\noindent {\bf Remark 4-2.} The assumption $H_{f\theta _k}^*=H_{ff}^*\rho _k$ can be relaxed to $H_{f\theta _k}^*=H_{ff}^{*\gamma }\rho _k$ ($0<\gamma <1$), which is an assumption sometimes made in the analysis of the kernel method \cite{Fukumizu2013}. In this case, the convergence rate should be suitably modified.

\noindent {\bf Remark 4-3.} Apart from the Neyman near-orthogonality, Chernozhukov et al. \cite{Chernozhukov2018} provided additional conditions on score regularity and quality of nuisance parameters (Assumptions 3.3 and 3.4 of their paper). Their proof of asymptotic unbiasedness of $\theta $ relies on these assumptions. In Appendix \ref{appendix_score_regularity}, we prove that these conditions are satisfied for suitably chosen $\alpha $ and $\beta $.

The above score functions may not be intuitively understood. Inspecting the concrete representations of $\partial _\theta \ell$, $\partial _f\ell$, $H_{\theta f}$ and $H_{ff}$, one can derive the following more intuitively understandable formula as well. 

\noindent {\bf Proposition 3 (Intuitive Neyman orthogonal scores for ML in our model).} {\it Assume that the positive Borel measures $\mu _a$ ($a\in \mathcal{A}$) defined as 
\begin{eqnarray}
\mu _a(E)\overset{\mathrm{def}}{=}\int _0^{\infty }\int _0^{C}\int _{E}P(A_{t}=a|X_t,T>t)dP(X_t|T>t)P(T>t)dtdP(C) 
\end{eqnarray}
$(E\in \mathfrak{M}(\mathcal{X}):$ Borel measurable sets$)$  are absolutely continuous to each other and have Radon-Nykodym derivatives 
\begin{eqnarray}
\frac{d\mu _k}{d\mu _0}=e^{g_k^*}, \label{definition_gk} 
\end{eqnarray}
for which $\mu _k$ denotes $\mu _a$ such that $a_k=1$, and $g_k^*$ is a measurable function on $\mathcal{X}$.  
Then, the score functions for estimating $\theta $,
\begin{eqnarray}
\phi _k(D; \theta , (f, g_k))&=&I_{[0,C]}(T)A_{k,T}e^{-\theta _k}(1+e^{-g_k(X_t)})-\int _{0}^{T\land C}A_{k,t}e^{f(X_{t})}(1+e^{-g_k(X_{t})})dt \nonumber \\
&&\hspace{-3.0cm}+\int _{0}^{T\land C}(1-\sum _{\ell \in \mathcal{K}}A_{\ell ,t})e^{f(X_{t})}(1+e^{g_k(X_t)})dt -I_{[0,C]}(T)(1-\sum _{\ell \in \mathcal{K}}A_{\ell,T})(1+e^{g_k(X_T)}), \label{ML_scores}
\end{eqnarray}
$(k\in \mathcal{K})$, are Neyman orthogonal. Here, $D$ denotes $(T, C, X, A)$.}

\begin{proof}
See Appendix \ref{appendix_prop3}. 
\end{proof}

\noindent {\bf Remark 5-1.} As one might notice, the crux of the construction of the above orthogonal scores lies in balancing the second and third terms of Eq.(\ref{ML_scores}) by $g_k$. Roughly speaking, for each value of covariates, $g_k$ weights the integrand with the inverse-treatment-probability for $A_k=1$ and $A=0$. Here, note that this treatment probability is marginalized with respect to time. 

\noindent {\bf Remark 5-2.} Note that the estimation of nuisance parameter $\{ g_k\} _{k\in \mathcal{K}}$ can be naturally formulated as the following logistic regression:  
\begin{eqnarray}
\min _{g_k}\widehat{E}\left [\int _{0}^{T\land C}\left \{ A_{k,t}\ln (1+e^{-g_k(X_{t})})+(1-\sum _{\ell \in \mathcal{K}} A_{\ell ,t})\ln (1+e^{g_k(X_{t})})\right \} dt \right ] +\zeta _{n,k}\| g_k\| _{\mathcal{H} _k}^2, \label{g_logistic}
\end{eqnarray}
for each $k\in \mathcal{K}$. Here, we have introduced a regularization hyperparameter $\zeta _{n,k}>0$.

\subsection{Extension to models with latent variables} \label{sec2_4}
Suppose that the Assumption 7 is violated due to the presence of unobserved factors affecting the outcome, and covariates cannot adjust the resultant temporal change in the risk set. In this case, as we see in the simulation study below, the inclusion of the time elapsed after enrollment in the set of covariates may improve the description of event occurrence. For such cases, if additional data about the unobserved factors cannot be acquired, the only alternative is to explicitly model the unobserved factors as latent variables. 

To simplify the argument and notation, let us assume that the unobserved factor (denoted by $W\in \{ 0, 1\} $) is a single time-independent binary variable (e.g., a variable indicating the presence or absence of a genetic risk factor). And suppose that we use a model with a latent variable $Z\in \{ 0, 1\} $ defined by the following prior distribution of $Z$ and $Z$-dependent conditional hazard:
\begin{eqnarray}
Q_{\beta }(Z|X_0,A_0)\overset{\mathrm{def}}{=}\frac{1}{1+\exp (-(2Z-1)(\sum _{j\in \mathcal{J}}\beta _jX_{j0}+\beta _0))}, \label{Z_prior}
\end{eqnarray}
and 
\begin{eqnarray}
\widetilde{h}(t|A_t, X_t, Z)\overset{\mathrm{def}}{=}\exp (\theta ^\prime A_t+f(X_t)+\kappa Z). \label{extended_hazard_model}
\end{eqnarray} 
The latter yields the following $Z$-dependent log-likelihood: 
\begin{eqnarray}
Q_{\widetilde{h}}(T|X,A,Z)=I_{[0,C]}(T)(\theta ^{\prime}A_T+\kappa Z+f(X_T))-\int _0^{T\land C} \exp(\theta ^{\prime }A_t+\kappa Z+f(X_t))dt. \label{Z_log_likelihood}
\end{eqnarray}

In the above, we have introduced parameters $\beta \in \mathbf{R} ^{\mathrm{dim} (X_0)+1}$ relating baseline values of covariates (at $t=0$) to the prior distribution of $Z$, and a parameter $\kappa \in \mathbf{R}$ representing the effect of the latent variable. 

The same argument for Proposition 1 applies to this model. Suppose that the data-generating process is described by 
\begin{eqnarray}
P(W|X_0,A_0)=\frac{1}{1+\exp (-(2W-1)(\sum _{j\in \mathcal{J}}\beta _j^{*}X_{j0}+\beta _0^*))}, \label{W_prior}
\end{eqnarray} 
and 
\begin{eqnarray}
\lim _{\delta t\rightarrow +0}\frac{P(I_{(t,t+\delta t]}(T)=1|T>t ,A_t=a, X_t=x, W=1)}{P(I_{(t,t+\delta t]}(T)=1|T>t ,A_t=a, X_t=x, W=0)} =const,
\end{eqnarray}
regardless of the values of $x$, $a$ and $t$. Then, replacing $X$ with the combination $(X, W)$ or $(X, Z)$ in Assumptions 1--8 and Proposition 1 and assuming that the model is identifiable, one obtains the same interpretation of hazard ratios.   

Then, one can apply Proposition 2 to derive Neyman-orthogonal scores from the following marginal log-likelihood for this model: 
\begin{eqnarray}
\sum _{i\in \mathcal{I}}\ln Q_{\widetilde{h}}(T_i|A_i,X_i) =\sum _{i\in \mathcal{I}}\max _{r_i}\sum _{Z_i\in \{ 0, 1\} } r_i(Z_i)\ln \frac{Q_{\widetilde{h} }(T_i|X_i, A_i, Z_i)Q_{\beta }(Z_i|X_i,A_i)}{r_i(Z_i)}, \label{log_marginal_likelihood}
\end{eqnarray}
with the aid of variational distributions $\{ r_i\} _{i\in \mathcal{I}}$ constrained by $r_i(0)+r_i(1)=1$. It should be noted that, in the application of Proposition 2, $f$ is replaced by $\mathrm{vec}(f,\kappa,\beta)$. For the above latent-variable model, a more intuitively understandable formula corresponding to Proposition 3 is not available.

\subsection{Design of models with multiple RKHSs} \label{sec2_5}
Although we have discussed a single RKHS for model $\mathcal{M}$, from a practical point of view, it is much more convenient to define $f$ in the following form \cite{Lanckriet2004, Suzuki2013}: 
\begin{eqnarray}
f(X)=\sum _{k_\ell \in \mathfrak{K}}f_{\ell }(X)+b, \ \ f_{\ell } \in \mathcal{H} _{k_\ell}, \ \ b\in \mathbf{R}. \label{MKL}
\end{eqnarray}
In the above, we have used a collection of RKHSs each of which is associated with a kernel function $k_\ell \in \mathfrak{K}$. With this formulation, one can separately model the effects of different factors, such as demographics, comorbidities and drug use. We provide examples in simulation studies below. Use of multiple kernels also allow us to construct $f$ by trial and error. One can observe whether incorporation or removal of a component function associated with a covariate improves the data description by calculating BME. On can also validate Assumptions 5--7 by checking whether incorporation of $f_{\ell }$ representing, for example, time inhomogeneity improves BME. Here, note that criteria for model selection must be BME (or Bayesian information criteria applicable to only parametric models), not cross-validation error, because the former has consistency in terms of model selection in the large sample limit, but the latter not. 

The theory developed in the previous sections apply to the above multiple-kernel model as well, because the sum of functions in different RKHSs is known to belong to an RKHS associated with a single composite kernel \cite{Aronszajn1950, Suzuki2013}. The squared norm of $f$ defined by this composite kernel is dominated by the square sum of the norms of the component functions \cite{Aronszajn1950}. Thus, the entire theory is not affected by the use of a multiple-kernel model. The theoretical properties of ML estimation with the model in Eq.(\ref{MKL}) under a 1-norm, 2-norm or mixed-norm regularization have been extensively studied \cite{Bach2008, Meier2009, Koltchinskii2010, Suzuki2013}. We employ the 2-norm-regularized version for which calculation of BME is tractable.  

\subsection{Estimation algorithm} \label{sec2_6}
\noindent In this section, we describe the concrete procedure for computing the debiased estimators of the hazard ratios for treatment. How the algorithm is numerically implemented is detailed in Appendix \ref{appendix_numerics}.  

\noindent {\bf Step 1: Model selection and determination of hyperparameters}

We first choose the best possible combination of RKHSs by trial and error by comparing BME in the manner we have described in the previous section. We also examine the validity of Assumptions 5--7 in this step. To compare BME for different models, we first perform a grid search of the optimal hyperparameter values for regularization and the bandwidth of kernels for each model, examining values with intervals of approximately $\ln(1.5)$ on the logarithmic scale (e.g. $2.0, 3.0, 5.0, 7.0, 10.0\ldots$) for each hyperparameter. Then, the BME values of different models calculated with the optimal hyperparameter values are compared. When we think that we have obtained the best possible model, we proceed to the next step. There is, however, a possibility that the validity of Assumptions 5--7 is denied for the best possible model at hand. In this case, one needs to review the study design, obtain new covariates or incorporate latent variables to the model.

\noindent {\bf Step 2: Estimation of nuisance parameters}

For the model chosen in Step 1, we calculate nuisance parameters. Following Chernozhukov et al. \cite{Chernozhukov2018}, we perform ``cross fitting'' as follows. To debias $\theta $, the estimation errors in $f$ and $H$ (or $g_k$) must be independent of the data $D_i$ in the argument of score functions. To achieve this, the entire dataset $D$ are first partitioned into $M$ groups of approximately equal size ($D^{(1)}\cdots D^{(M)})$. We perform the estimation of nuisance parameters $M$ times, using $D_{\mathrm{train}}^{(m)}\overset{\mathrm{def}}{=}D\backslash (D^{(m)}\cup D^{(m+1\ (\mathrm{mod} M))})$ as a training set, validating the estimation result with $D_{\mathrm{val}}^{(m)}\overset{\mathrm{def}}{=}D^{(m+1\ (\mathrm{mod} M))}$ and then yielding the values used for score functions of $D_{\mathrm{hout}}^{(m)}\overset{\mathrm{def}}{=}D^{(m)}$ for $1\leq m\leq M$. 

Concretely, we first perform ML estimation to obtain $\widehat{\theta ^{(\backslash m)}}$ and $\widehat{f^{(\backslash m)}}$ as the minimizer of the negative log-likelihood of $D_{\mathrm{train}}^{(m)}$ using the optimal hyperparameter values obtained in Step 1. Then, we proceed to the computation of additional nuisance parameters $\widehat{H}$ (or $\{ \widehat{g}_k\} _k$) based on $D_{\mathrm{train}}^{(m)}$. In this step, we also need to identify the value of the regularization hyperparameter $\zeta _n$ for these nuisance parameter. For the determination of $\zeta _n$ for $\widehat{H}$, we examine the following cross-validation error 
\begin{eqnarray}
\mathrm{CVErr} _{H} \overset{\mathrm{def}}{=} \sum _{k\in \mathcal{K}}\sum _{1\leq m\leq M} \left \| \widehat{H}_{\theta_kf}^{(m,\mathrm{val})}-\widehat{H} _{\theta _kf}^{(\backslash m)}(\widehat{H}_{ff}^{(\backslash m)}+\zeta _n)^{-1}\widehat{H} _{ff}^{(m,\mathrm{val})}\right \| _{\mathcal{H}_k}^2 \label{CVErrH}
\end{eqnarray}
Here, the $\widehat{H}_{\theta f}^{(m,\mathrm{val})}$ and $\widehat{H} _{ff}^{(m,\mathrm{val})}$ are the block components of the Hessian of the negative log-likelihood of $D_{\mathrm{val}}^{(m)}$.

For the determination of $\{\zeta _{n,k}\} _k$ for $\{ g_k\} _k$, the logarithmic loss in Eq.(\ref{g_logistic}) may underestimate the error due to large $e^{g_k}$ and $e^{-g_k}$ in the score functions. We therefore suggest computing the following cross-validation error: 
\begin{eqnarray}
&&\hspace{-1.5cm}\mathrm{CVErr} _{g_k}\overset{\mathrm{def}}{=}\nonumber \\
&&\hspace{-1.5cm}\sum _{1\leq m\leq M} \left [ \sum _{i\in D_{\mathrm{val}}^{(m)}}\int _{0}^{T_i\land C_i}\left \{ A_{ik,t} (e^{-\widehat{g_{k}^{(\backslash m)}}(X_{i,t})}-1)+(1-\sum _{\ell \in \mathcal{K}} A_{i\ell ,t})(e^{\widehat{g_{k}^{(\backslash m)}}(X_{i,t})}-1)\right \} dt\right ]^2. \label{CVErrg}
\end{eqnarray}
The above error design exploits the fact that, for $\widehat{P}(A_k=1|X)\rightarrow P(A_k=1|X)$, 
\begin{eqnarray}
P(A_k=1|X)\frac{1}{\widehat{P}(A_k=1|X)} +(1-P(A_k=1|X))\frac{1}{1-\widehat{P} (A_k=1|X)} \rightarrow 2.
\end{eqnarray}
Here, the trivial solution $\widehat{P}(A_k=1|X)=\frac{1}{2}$ also minimizes $\mathrm{CVErr} _{g_k}$ and should be avoided by checking BME of the logistic regression (Eq.(\ref{g_logistic})) as well.

\noindent {\bf Step 3: Debiased estimation of $\theta $ and its standard error. } \\
The debiased estimator $\overline{\theta }$ of $\theta ^*$ is obtained as the zero of the following:
\begin{eqnarray}
\sum _{1\leq m\leq M}\sum _{i\in D_{\mathrm{hout}}^{(m)}} \phi (D_i; \theta ,\widehat{\eta } ^{(\backslash m)}),
\end{eqnarray}
with $\widehat{\eta }^{(\backslash m)}=(\widehat{f}^{(\backslash m)}, \widehat{H}^{(\backslash m)})$ (or $(\widehat{f}^{(\backslash m)}, \widehat{g} _k^{(\backslash m)}$)). According to Chernozhukov et al. \cite{Chernozhukov2018}, the asymptotic standard error of the estimator is given by $\Sigma ^{-1/2}(\overline{\theta }-\theta ^*)\overset{d}{\rightarrow } \mathrm{Normal} (0,1)$ with
\begin{eqnarray}
\widehat{\Sigma } &\overset{\mathrm{def}}{=} &\widehat{J} ^{-1}\left (\frac{1}{n^2}\sum _{1\leq m\leq M}\sum _{i\in D^{(m)}} \phi (D_i; \overline{\theta } ,(\widehat{f} ^{(\backslash m)}, \widehat{\eta} ^{(\backslash m)}))\phi (D_i; \overline{\theta } ,(\widehat{f} ^{(\backslash m)}, \widehat{\eta} ^{(\backslash m)}))^{\prime }\right ) \widehat{J} ^{-1\prime }, \\
\widehat{J} &\overset{\mathrm{def}}{=} &\frac{1}{n}\sum _{1\leq m\leq M}\sum _{i\in D^{(m)}} \frac{\partial }{\partial \theta }\phi (D_i; \theta ,(\widehat{f} ^{(\backslash m)}, \widehat{\eta} ^{(\backslash m)})) |_{\theta =\overline{\theta }}.
\end{eqnarray}
Thus, in the simulation study described below, the square roots of the diagonal elements of $\widehat{\Sigma }$ serve to scale the errors from the true value to obtain the {\it t}-statistics. 

\noindent {\bf Remark 6.} The orthogonal scores as functions of $\theta $ are given in Proposition 3 for the case with $g_k$. For the case with $H$, a procedure for obtaining functions of $\theta $ is given in Appendix \ref{appendix_numerics}.   

\section{Results of Numerical Simulations}  \label{sec3}

\subsection{Simulation result 1: Adjustment for observed confounders} \label{sec3_1}
This section presents a simulation study designed to demonstrate that, in a clinically plausible setting, the proposed method estimates hazard ratios for treatment with minimal bias. Suppose we randomly sample subjects from an electronic medical record that contains the medical history of a large local population. Suppose further that 2,000 subjects aged 50--75 years enrolled at uniformly random times between 2000 and 2005. The chosen subjects are followed up for $C\overset{i.i.d.}{\sim } \mathrm{Uniform}[5, 10]$ years, with timely recording of the onset of comorbidities, the administration of drugs and the occurrence of a disorder defined as an outcome event. 

As a typical setting in which confounders bias the estimation of the treatment effect, suppose that subjects are newly diagnosed with condition 1 at a constant rate (2.5\%/month), which may be a prodrome for the outcome event. A drug that is suspected to be a risk factor for an outcome event may be used to treat this condition. In the simulation, the probability per month of initiating this drug is $0.004+0.2X_{1t}\exp (-X_{1t}/0.6)/(0.6)^2$ with a covariate $X_{1t}$ [year] indicating the duration of suffering condition 1 before time $t$, and the probability per month of discontinuing the drug is 0.01. Use of this drug increases the risk of developing condition 2. The probability per month of developing condition 2 at time $t$ is $0.05+0.05\int _{-\infty }^{t}B_{s}\Theta (t-s-1/12)e^{-3(t-s-1/12)}ds$, where $B_s$ takes a value of 1 if the subject is a current user of the drug at time $s$, or it takes a value of 0 otherwise. The above integral including the Heaviside step function $\Theta $ describes a short-term delayed increase in the risk of developing condition 2 subsequent to the drug use. 

Finally, suppose that the occurrence of the outcome event depends on both treatment and conditions 1 and 2, and its true risk per month is $\exp(-7.0+\theta _1^*A_{1t}+\theta _2^*A_{2t}+0.04\mathrm{age}_{t}+0.2\sin (\pi \mathrm{date}_{t}/8)-0.2\cos (\pi \mathrm{date}_{t}/6)+2.0X_1\exp (-X_{1t}/1.5)+P_2(1-\exp (-X_{2t}/2.5)))/12.0$, with $\theta _1^*=1.0$ and $\theta _2^*=2.0$, where we define two treatment variables $A_{1t}$ and $A_{2t}$, which take a value of 1 if and only if the subject use the drug for a period $\leq18$ and $>18$ months before time $t$, respectively. $P_2$ in the equation is a parameter for which we use different values. The covariates $\mathrm{age}_{t}$, $\mathrm{date}_{t}$ and $X_{2t}$ (duration of suffering condition 2) are measured in years.  

For this dataset, we performed debiased ML estimation using multiple-kernel models for $f$. Concretely, we used the following model: 
\begin{eqnarray}
f(X_t)=f_{\mathrm{age}}(\mathrm{age} _t)+f_{\mathrm{date}}(\mathrm{date} _t)+f_1(X_{1t})+ f_2(X_{2t})+b,
\end{eqnarray}
with $f_{\mathrm{age}}\in \mathcal{H} _{k_{\mathrm{age}}}$, $f_{\mathrm{date}}\in \mathcal{H} _{k_{\mathrm{date}}}$, $f_{1}\in \mathcal{H} _{k_{1}}$ and $f_{2}\in \mathcal{H} _{k_{2}}$. To correctly specify the model, we chose $k_{\mathrm{age}}$ to be a linear kernel and $k_{\mathrm{date}}$, $k_1$ and $k_2$ to be one-dimensional Gaussian kernels [e.g. $k_1(X_{i_1,t_1}, X_{i_2,t_2})=\exp (-(X_{i_11,t_1}-X_{i_21,t_2})^2/\sigma ^2)$] with bandwidth hyperparameter, $\sigma $. Although we can {\it a priori} identify the set of necessary kernels, one can find the appropriate set of kernels by trial and error, using BME as a criterion. In this step, we can also validate the model assumptions (Assumption 5--7) by examining whether the introduction of an additional kernel improves BME. We briefly illustrate the idea through an example. 

First, as we have described in the above, covariate $X_{2t}$ is necessary for predicting the onset of outcome events. This can be seen by comparing BME values for the correctly specified model and a misspecified model built by removing $f_2(X_{2t})$ and $\mathcal{H} _{k_2}$ from it (Fig.1(A)). The question naturally arising here is how one can see that there is a room for improvement when they have the misspecified model at hand. We propose to examine the assumption of time homogeneity (Assumption 7) as an indicator. In our example, comparison of BME values for the second, misspecified model with or without inclusion of the time elapsed after enrollment as a covariate shows the violation of the assumption, depending on the value of $P_2$ (Fig.1(B)), although the sensitivity is not high. The risk variation due to the time elapsed after enrollment indicates a temporal change in the risk set that covariates at hand do not account for, and therefore the model is missing some necessary factors (see the argument in Remark 3-1). For a correctly specified model, such violation of the assumption is not detected (Fig.1(C)). The violation of the time-homogeneity assumption is observed for models misspecified in different manners (which we omit to avoid redundancy). 

Next, suppose that we successfully identified the correctly specified model and proceeded to the calculation of debiased estimators of hazard ratios. We carried out this calculation in two different ways based on either one of Propositions 2 and 3. In both applications, we followed the procedure described in section 2.6. Additionally, in the application of Proposition 3, we use the following multiple-kernel model for the estimation of $g_\ell $ ($\ell \in \mathcal{K}$):
\begin{eqnarray}
g_\ell (X_t)=g_{\ell, \mathrm{age}}(\mathrm{age} _t)+g_{\ell ,\mathrm{date}}(\mathrm{date} _t)+g_{\ell 1}(X_{1t})+ g_{\ell 2}(X_{2t})+b,
\end{eqnarray}
with $g_{\ell, \mathrm{age}}\in \mathcal{H} _{k_{\mathrm{age}}}$, $g_{\ell \mathrm{date}}\in \mathcal{H} _{k_{\mathrm{date}}}$, $g_{\ell 1}\in \mathcal{H} _{k_{1}}$ and $g_{\ell 2}\in \mathcal{H} _{k_2}$, where all RKHSs are associated with a one-dimensional Gaussian kernel. On top of the kernels described above, we also examined whether the inclusion of two-dimensional Gaussian kernels such as $k_{12}(X_{i_1,t_1}, X_{i_2,t_2})=\exp (-((X_{i_11,t_1}-X_{i_21,t_2})^2+(X_{i_12,t_1}-X_{i_22,t_2})^2)/\sigma _2^2),$ improves BME for the logistic regression in Eq.(\ref{g_logistic}). In the present case, the two-dimensional kernels did not improve BME. 

In the tuning of hyperparameter values for nuisance parameters $H$ (or $g_\ell $), we examined how the hyperparameters affect $\mathrm{CVErr} _H$ (Fig.1(D)) and $\mathrm{CVErr}_{g_\ell}$ (Fig.1(F)), respectively. In the application of Proposition 3, BME for the logistic regression and $\mathrm{CVErr}_{g_\ell}$ identified different optimal values (Fig.1(E) and (F)). 

After taking all of the steps in the Algorithm, we calculated debiased estimate $\overline{\theta }$ using the score functions obtained with either $(H_{\theta f}, H_{ff})$ or $g_1$. To measure the bias in the estimator, we obtained estimates with 500 datasets generated from the identical process described above with different seeds for the random-number generator. The {\it t}-statistics ($\overset{\mathrm{def}}{=}(\widehat{\theta }_k-\theta _k^*)/\widehat{\mathrm{SE} _{\theta _k}}$) in Fig.1(G) and (H) shows that the debiased estimator identified the true value with minimal bias, in contrast with the naive ML estimator. In Fig.1(H), we also observed that $g_1$ tuned with $\mathrm{CVErr}_{g_1}$ debias the estimator more effectively than $g_1$ tuned with BME. 

\begin{figure}[!ht]
\centering
\includegraphics[width=135mm]{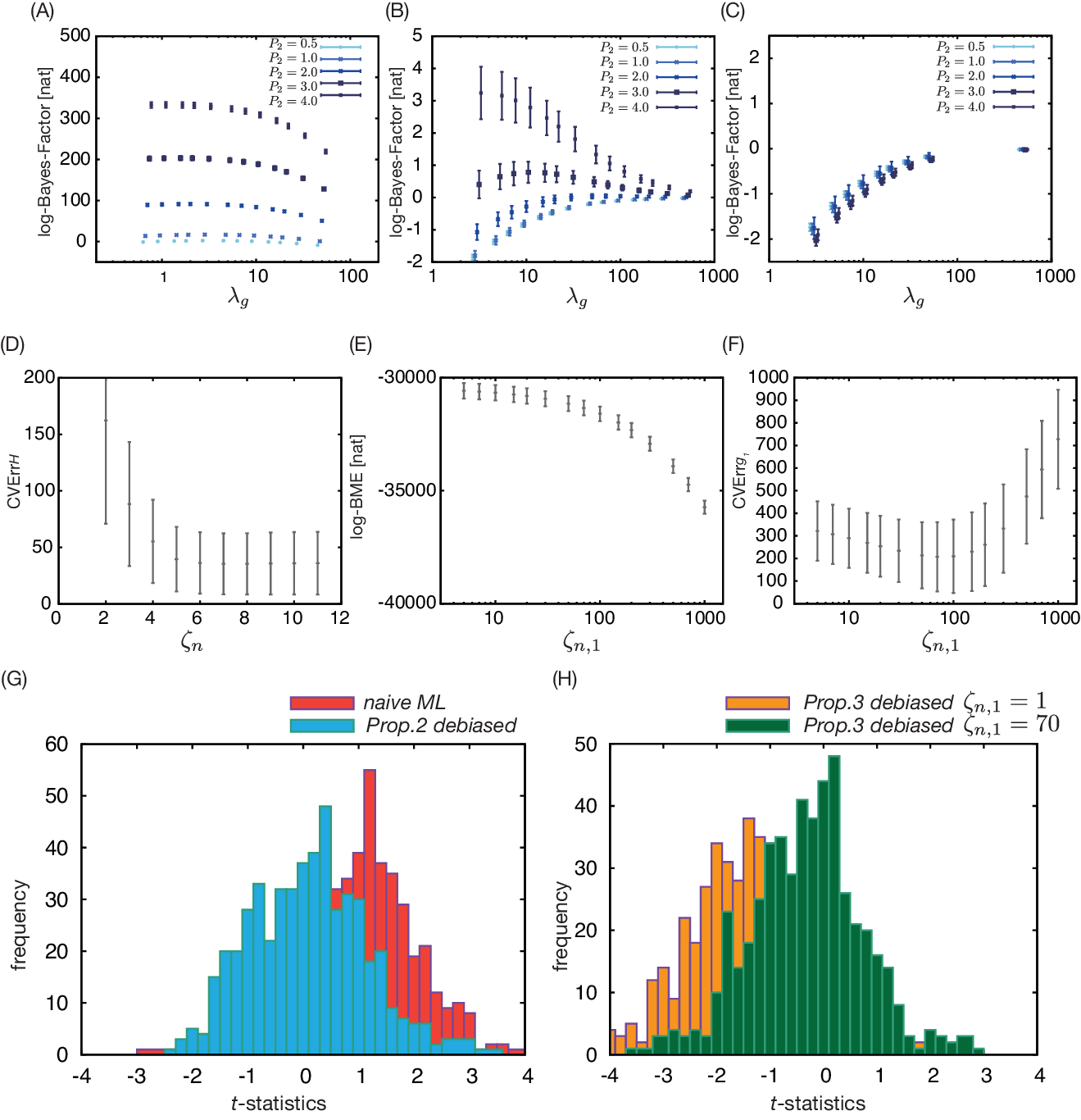}
\caption{Numerical results for a clinically plausible data-generation process. (A)--(C): The mean and standard error of the natural logarithms of the Bayes factors for (A) the correctly specified model vs the misspecified models lacking $f_2(X_{2t})$, (B) the misspecified models with vs without the inclusion of $f_t(t)$ and (C) the correctly specified models with vs without the inclusion of $f_t(t)$, calculated with ten bootstrap datasets and different regularization parameters for Gaussian kernels $\lambda _g$. The other hyperparameters are fixed to the optimal values for the smaller model. The log-Bayes factors are calculated by subtracting the log-BME of the larger model from that of the smaller model. (D)--(F) The mean and standard error of (D) $\mathrm{CVErr} _H$, (E) log-BME for $g_1$ and (F) $\mathrm{CVErr}_{g_1}$ calculated with ten bootstrap datasets and different regularization parameters. (G) and (H): The histograms of {\it t}-statistics measuring the bias in the naive ML estimator of $\theta _1^*$ and the debiased ML estimators of $\theta _1^*$ based on Propositions 2 and 3. For the estimator based on Proposition 3, results with nuisance parameter $g_1$ estimated with $\zeta _{n,1}=70$ and $\zeta _{n,1}=1$ are shown.}
\label{fig1}
\end{figure}

\subsection{Simulation result 2: Estimation of treatment effect in population with heterogeneous risk} \label{sec3_2}
Suppose that most of the simulation settings are the same as the previous one, but the enrolled subjects are divided into two distinct risk groups denoted by $W\in \{0,1\}$, and this is reflected in a baseline covariate value. More precisely, the number of subjects, their age, enrollment date, censoring, onset probabilities of conditions 1 and 2 and treatment assignment are the same as the previous one. Suppose, however, that blood-test results $X_{j0}^{(\mathrm{test})}\overset{i.i.d.}{\sim } \mathrm{Normal} (0, \sigma _j^{(\mathrm{test})2})$ at baseline ($t=0$) are available ($1\leq j\leq 3$), and the risk group to which the subject belongs is probabilistically related to $\{ X_{j0}^{(\mathrm{test})}\} _j$ as
\begin{eqnarray}
P(W|A,X)=\frac{1}{1+\exp (-(2W-1)(\sum _{j}\beta _j^{*\prime }X_{j0}^{(\mathrm{test})}+\beta _0^*))}, \label{W_prior_again}
\end{eqnarray}
with $\{ \sigma _j^{(\mathrm{test})}\} _{1\leq j\leq 3}=\{ 2.0, 1.0, 4.0\} $ and $\{ \beta ^*_j\} _{0\leq j\leq 3}=\{ -0.5, 1.0, 0.0, 0.0\} $. With variable $W$, the risk of outcome event per month is given by $\exp(-7.0+\theta _1^*A_{1t}+\theta _2^*A_{2t}+0.04\mathrm{age}_{t}+0.2\sin (\pi \mathrm{date}_{t}/8)-0.2\cos (\pi \mathrm{date}_{t}/6)+2.0X_1\exp (-X_{1t}/1.5)+P_2(1-\exp(-X_{2t}/2.5))+\kappa ^*W)/12.0$ with $\theta _1^*=1.0$, $\theta _2^*=2.0$, $P_2=0.5$ and $\kappa ^*=1.0, 2.0$ or $3.0$.

We first performed analysis with only observed covariates $(X_t, X_t^{(\mathrm{test})})$, not using latent variables. Here, we assumed $X_t^{(\mathrm{test})}=X_0^{(\mathrm{test})}$. In the validation of assumptions, the inclusion of the time elapsed after enrollment in the set of covariates improved BME (Fig.2(A)), suggesting that the enrolled subjects are heterogeneous, presumably due to an unobserved factor. Then, we analyzed the dataset with a latent variable, maximizing the marginal log-likelihood with the EM algorithm (see e.g., Chapter 9 of Ref.\cite{Bishop2006}), which led to further improvement of BME (Fig.2(B)). In this calculation, we observed that the optimization becomes numerically unstable for small $\kappa (\leq 1.0)$ presumably because of the singularity of the model. We examined the estimated values of $\theta $ with 500 datasets generated by the identical process with different seeds for its random-number generator, and the results confirmed that the suggested debiased estimators identify the true value with minimal bias, in contrast with the naive ML estimator (Fig.2(C)).

\begin{figure}[!ht]
\centering
\includegraphics[width=120mm]{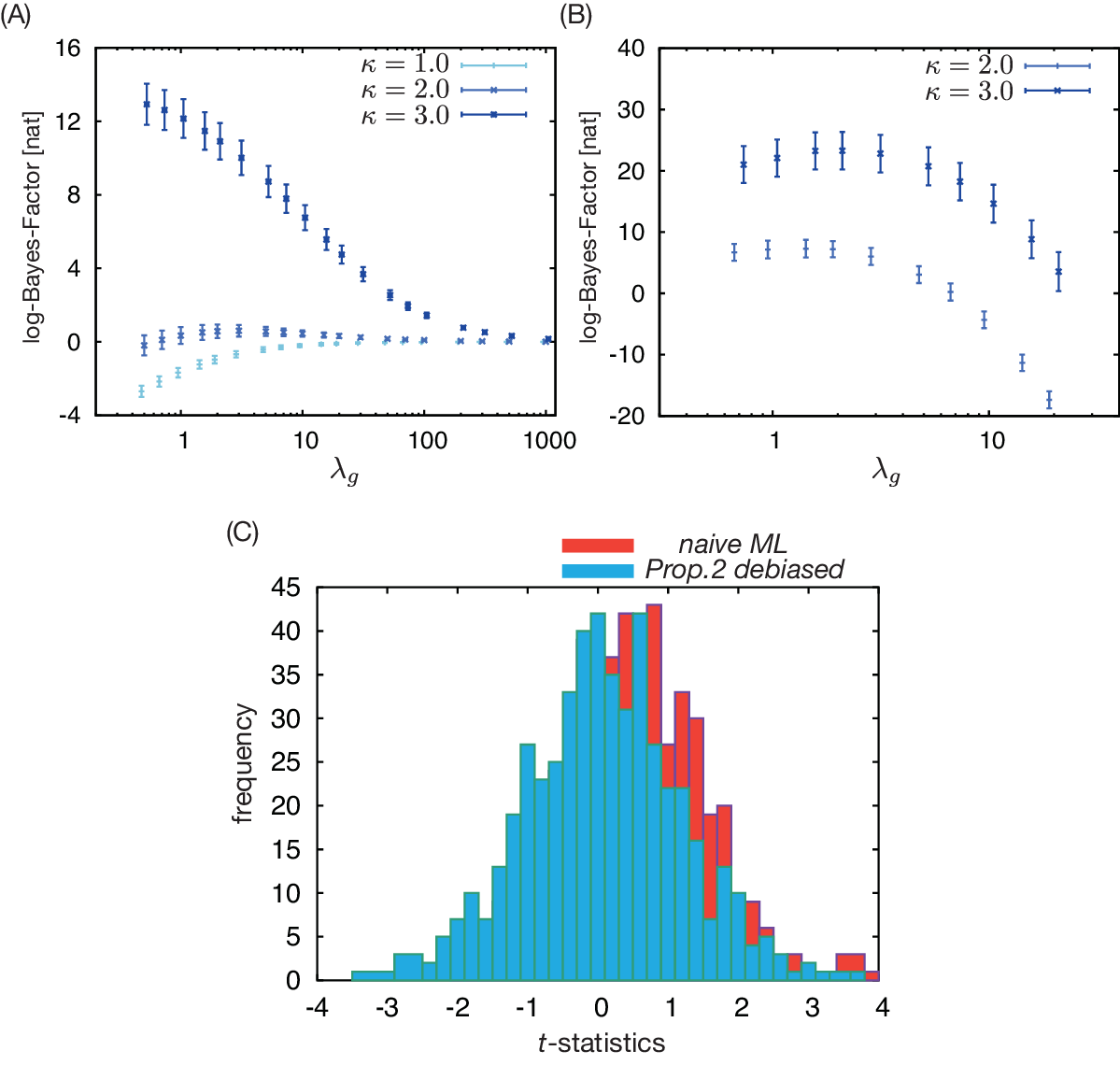}
\caption{Numerical results for a clinically plausible data-generation process with a time-independent unobserved factor affecting outcome. (A) and (B): The mean and standard errors of log-Bayes factors for (A) the misspecified models with only the observed covariates with vs without the inclusion of $f(X_{1t}^{(\mathrm{test})}, t)$ and (B) the correctly specified model with a latent variable vs the misspecified model with only the observed covariates, but without the inclusion of $f(X_{1t}^{(\mathrm{test})}, t)$, calculated with ten bootstrap datasets and different regularization parameters for Gaussian kernels. The other hyperparameters are fixed to the optimal values for the smaller (null) model. (C) The histograms of {\it t}-statistics measuring the bias in the naive ML estimator of $\theta _1$ and the debiased ML estimator of $\theta _1$ based on the correctly specified model with a latent variable.}
\label{fig2}
\end{figure}

\section{Discussion} \label{sec4}

Preceding studies \cite{Hernan2010, Aalen2015, Martinussen2020} showed that the Cox model allows multiple, contradictory interpretations due to the selection process in the risk set. To alleviate this difficulty, we proposed a new framework that combines an exponential hazard model with machine learning models. With the observation that the baseline hazard of the Cox model is a major source of the uninterpretability, we constructed the model without a baseline hazard. For this model, with a few testable assumptions as well as conventional assumptions for causal inference in observational studies, we clarified the context in which the estimated hazard ratios are causally interpreted. In this argument we have demonstrated that the source of uninterpretability is mostly removed (Remark 3-1). Then, we developed a framework to systematically seek for the best possible model with the aid of machine learning and then compute debiased estimates of hazard ratios with the obtained model. Numerical simulations demonstrated that wrong description of the selection process in the risk set can be detected as violation of a time-homogeneity assumption in this framework. In the simulation, it was also demonstrated that our theoretically justified debiased ML estimators of hazard ratios identify the true values with minimal bias.

Epidemiologists and statisticians might find it psychologically difficult to abandon the Cox models with baseline hazard because of its established role as a default model. The popularity of the Cox model is due to its semiparametric nature. Leaving the aspects of data difficult to model as a blackbox is an attractive idea. Furthermore, the Cox's MPL estimator \cite{Cox1972} for log-hazard ratios of treatment groups is asymptotically efficient \cite{Efron1977, Oakes1977}, which is a natural consequence of obviating the estimation of the baseline hazard. Because of this asymptotic efficiency, all other estimators, including the ML estimator, have essentially been reduced to theoretical subjects, while a relatively small number of studies have suggested the advantage of using the others (e.g. \cite{Ren2011, Thackham2020}). The present results suggest the need to reconsider this situation by reevaluating the tradeoff between the efficiency and uninterpretability due to the unspecified baseline hazard. The disadvantage of using MPL is not only the uninterpretability. First of all, the baseline hazard is impossible to interpret from the physical and biological points of view. No physical process depends on the time elapsed after the enrollment. Thus, its use is solely justified by convenience. However, from a technical point of view, comparing the estimation of $g_k$ in the present study and the inverse-probability-of-treatment weight for marginal structural Cox models used for a similar purpose \cite{Hernan2001}, one can see that the former is a marginal estimand in terms of time and hence usually easier to estimate. The standard method for the latter, that is, estimating the transition probability of treatment and censoring for each value of covariates and each time $t$, possibly works better in a simple model setting, such as analysis involving only a few--several discrete variables. However, carrying this out under the nonlinear effect of a high-dimensional covariate is prohibitive. In fact, the recently proposed application of doubly robust estimation to marginal structural Cox models are still limited to simple model settings to which machine learning cannot be incorporated \cite{Luo2024}. This is why we could not directly compare the results with our method and those with the marginal structural Cox model. Another negative factor for MPL approach is that the combination of Bayes theorem and partial likelihood is only approximately justifiable, so extending Cox regression to Bayesian settings is not straightforward, although efforts in this direction have been made for Cox models as well \cite{Zhang2023}. Considering these negative factors for the MPL approach, we conclude that the debiased ML approach should be reconsidered as an option in causal inference for observational studies with uncontrolled dynamic treatment and real-time measurement of a large number of covariates.

It is, however, fair to note that the applicability of our approach at the current stage is largely limited because Assumptions 1--8 must be satisfied for the application. Among these assumptions, Assumptions 1--4 are technical assumptions that are explicitly or implicitly made in the conventional approach with the marginal structural Cox model \cite{Hernan2001}. In particular, in causal inference with observational data, Assumption 2 stating conditionally randomized assignment of treatment for each value of covariate is thought to be crucial. If this does not hold, an experiment (intervention) is needed. See Remarks 1-1--1-4 for the reasoning of the other assumptions. Assumptions 5--8 essentially concern specification of a model. Assumption 6 is potentially removable. Estimation of heterogeneous treatment effect (i.e., covariate-dependent treatment effect) is a subject of active research \cite{Inoue2024} and an approach that adjusts the heterogeneity in the treatment effect in an unbiased manner may be incorporated into our approach. Then, the remaining problem is whether one can design a suitable set of variables for $X$ and $A$ (Assumption 5) and a correctly specified machine-learning model (Assumption 8) without sacrificing the assumption of time-homogeneity (Assumption 7). This will be achieved, if one can construct a model that simulates the data-generating process precisely, because the biophysical processes generating data do not depend on specific time points of observation. At the current stage, the relatively simple model we have used in this study is likely to suffer misspecification in real-world epidemiological problems. Therefore, our proposals stated in this study should be considered as a suggestion of future efforts to overcome this misspecification. Taking the current success of deep learning in real-world applications into account, we would like to suggest an optimistic view that models and data of a large scale in combination with Bayesian inference may remove any such misspecification problem. Before such a goal is achieved, no one knows a priori which of our approach and the Cox's MPL approach yields a value closer to the true hazard ratio for a given dataset.

In technical aspects, we indeed restricted our study to a simple setting for the sake of clear exposition. In this study, only non-informative right censoring is considered, only a single binary treatment variable can take the value of one, and only a single time-independent latent variable is allowed. However, the advantage of ML estimation has been demonstrated in the context of informative censoring \cite{Ma2014}. Therefore, extending our framework in this direction seems promising. The use of multiple independently changing treatment variables is used for the sake of an exposition, and our theories apply to general cases. Application of our framework to models with continuous treatment variables is also straightforward. Proposition 2 applies to models with linear functions of continuous treatment variables, while Proposition 3 cannot be used for this case. Since our theory and the framework of DML rely on the low dimensionality of treatment variables, extension to models with nonlinear treatment effect of continuous variables is beyond the fundamental limit of the current technique. However, use of multiple binary variables can approximately describe nonlinear dependence on treatment variables. This is the reason why we chose the binary setting. The extension of our framework to models with multiple (possibly continuous, time-dependent) latent variables would face the identifiability problem. Although progress has been made in both theoretical analysis of identifiability \cite{Allman2009, Allman2015, Gassiat2016a, Gassiat2016b} and numerical examination of identifiability in the context of biological modeling \cite{Wieland2021}, there is still a technical challenge. Related to this topic, using a singular model can lead to effective breakdown of asymptotic normality of estimators \cite{Watanabe2009}. In this case, the framework of debiased machine learning would itself need modifications, and the Laplace approximation of BME also fails and needs to be replaced by Monte-Carlo integration \cite{Calderhead2009} (or information criteria for singular parametric models \cite{Watanabe2013, Drton2017}). The above difficulties being mentioned, use of rich classes of latent-variable models for which inference methods have been extensively developed in machine learning (especially those for time-series data \cite{Moral2004, Chopin2020}) is yet an attractive, worthy challenge. In this challenge, our technique for debiasing will be useful.

\section*{Acknowledgements}
This research was supported by the Ministry of Education, Culture, Sports, Sciences and Technology (MEXT) of the Japanese government and the Japan Agency for Medical Research and Development (AMED) under grant numbers JP18km0605001 and JP223fa627011. We would like to express our gratitudes to two medical IT companies, 4DIN Ltd. (Tokyo, Japan) and Phenogen Medical Corporation (Tokyo, Japan) for financial support, while both companies had no role in the research design, analysis, data collection, interpretation of data, and review of the manuscript, and no honoraria or payments were made for authorship. 
\section*{Conflicts of Interest}
The authors declare no conflicts of interest. 
\section*{Author Contributions}
T.H. conceptualized the study. T.H. performed all of the mathematical works in the study. T.H. and S.A. designed models and simulation data. T.H. developed all program codes. T.H. and S.A. interpreted the analyzed results. T.H. wrote the manuscript. T.H. and S.A. have read and agreed to its contents and have approved the final manuscript for submissions.
\section*{Data Availability}
All of the source codes that support the findings of this study are available as supplementary materials. This will be publicized upon acceptance to a journal.

\section*{Appendix}

\begin{table}[H] 
\caption{Mathmatical notations\label{table1}}
\begin{tabular}{|l|l|}
\hline
{\bf Symbols}& {\bf Description}  \\
\hline \hline 
$\mathbf{N}$, $\mathbf{R} , \mathbf{R} ^d$ & The sets of natural and real numbers and $d$-dimensional Euclidean space ($d\in \mathbf{N}$) \\
$\mathcal{X} _1\times \mathcal{X} _2$ & Product set (or product space)\\
$(\mathcal{X} _1\times \mathcal{X} _2, \mathfrak{M}_{\mathcal{X} _1}\times \mathfrak{M}_{\mathcal{X} _2})$ & Measure space with product space and product $\sigma $-algebra \\
$v_1 \otimes v_2$ & Tensor product \\
$v^{\prime }$, $A^{\prime }$ & Transposition of vector $v$ and matrix $A$ \\
$a\land b $ & The smaller of $a$ and $b \in \mathbf{R}$   \\
$a\in \mathcal{A}$ & Element $a$ of a set $\mathcal{A}$ \\
$X\subset Y$ & Set inclusion (with possible equality) \\ 
$X\cup Y$, $X\cap Y$ & Union and intersection of sets \\
$|\mathcal{A}|$ & The number of elements in a set \\
$A \overset{p}{\rightarrow } $ B, $A \overset{d}{\rightarrow } B$ & Convergence in probability and in distribution \\
$b_n\sim o(a_n)$, $O(a_n)$ & Asymptotic notations for $n\rightarrow \infty $ (see Ref.\cite{Janson2011})  \\
$b_n\sim o_p(a_n)$, $O_p(a_n)$ & Probabilistic asymptotic notations for $n\rightarrow \infty $ (see Ref.\cite{Janson2011})  \\
($\arg$)$\min _{x\in \mathcal{X}}f(x)$ & (Element yielding) minimum of $f(x)$ over $\mathcal{X}$\\
($\arg$)$\max _{x\in \mathcal{X}}f(x)$ & (Element yielding) maximum of $f(x)$ over $\mathcal{X}$\\
$\sup _{x\in \mathcal{X}}f(x)$ & Supremum of $f(x)$ over $\mathcal{X}$\\
$E[\cdot ]$ ($E_{X\sim p}[\cdot ]$) & Expectation of the argument random variable (for the specified distribution)\\ 
$\overset{\mathrm{def}}{=}$ & Equation defining the object on the left-hand side \\
$\| v\| _2$, $\| v\| _{\mathcal{H}}$ & The norm of the metric vector space and the normed vector space $\mathcal{H}$\\
$\| A\| $ & The operator norm of a linear operator $A$ \\ 
$A \mathop{\perp\!\!\!\!\perp}  B$$(|C_1,\ldots )$ & Independence between $A$ and $B$ (conditioned on $C_1,\ldots $). \\
$i.i.d.$, ($\overset{i.i.d.}{\sim }$) & Independently and identically distributed (objects drawn from the right-hand side)\\  
$\int _Ef(X)d\mu (X)$ & Integration of $f(X)$ over the set $E$ with respect to the measure $\mu $ on the space for $X$ \\
$\mathrm{vec}(v_1, v_2\ldots v_k)$ & Natural mapping of vectors into their product space \\
$\partial _{v}(\cdot) $ & Abbreviation of partial derivative $\frac{\partial }{\partial v} (\cdot) $ \\
$\partial _{v}(\cdot) [v_1]|_{v_2}$ & Gateaux derivative $\lim _{\epsilon \rightarrow +0}\epsilon ^{-1}((\cdot )(v_2+\epsilon v_1)-(\cdot)(v_2))$ \\
$\mathrm{Uniform} [a,b]$ & Uniform probability distribution over the interval $[a,b]$ \\
$\mathrm{Normal} (\mu ,\Sigma)$ & Gaussian probability distribution with mean $\mu$ and (co)variance $\Sigma$ \\ 
\hline
\end{tabular}
\end{table}

\section{Numerical implementation of inference with multiple-kernel models} \label{appendix_numerics}
Regularized ML estimation of $\theta (\in \mathbf{R})$ and $f (\in \mathcal{H}$ : RKHS for the composite kernel determined by Eq.(\ref{MKL})) with discretized timesteps ($\mathfrak{T} _i=\{0, \Delta t, 2\Delta t, \cdots (\leq C_i\land T_i)\}$ for subject $i$) under a 2-norm regularization is formulated as
\begin{eqnarray}
\min _{\theta , f}\sum _{i\in \mathcal{I}}\sum _{t\in \mathfrak{T} _i}U(D_{i,t}; \theta, f)  +\sum _{k\in \mathfrak{K}}\frac{\lambda _k}{2} \alpha _k^{\prime }G_k\alpha _k,  \label{min_problem}
\end{eqnarray} 
where the Gram matrix, $(G_k)_{it, i^{\prime }t^{\prime }}=k(X_{i,t}, X_{i^{\prime }, t^{\prime }})$ relates $f$ and $\alpha _k$ via 
\begin{eqnarray}
f(X_{i,t})=\sum _{k\in \mathfrak{K}}\sum _{i^{\prime }t^{\prime }}(G_{k})_{it, i^{\prime }t^{\prime }}\alpha _{k, i^{\prime }t^{\prime }}+b
\end{eqnarray}
together with a bias parameter $b\in \mathbf{R}$, and 
\begin{eqnarray}
U(D_{i,t}; \theta , f)=-I_{[t, t+\Delta t)}(T_i)(\theta ^{\prime }A_{it}+f(X_{it}))+\exp (\theta ^{\prime }A_{it}+f(X_{it}))
\end{eqnarray}
 denotes the timestep-wise negative likelihood function. Note that the integration stepwidth $\Delta t$ can be omitted in the representation of $U$ because use of different values of $\Delta t$ (namely change of units) affects the estimate of only $b$. Concretely, we use linear or 1--3 dimensional Gaussian kernels, such as $k(X_{i_1,t_1}, X_{i_2,t_2})=X_{i_1j,t_1}X_{i_2j, t_2}$ and $k(X_{i_1,t_1}, X_{i_1, t_1})=\exp (-\sum _{1\leq \ell \leq d}(X_{i_1j_{\ell },t_1}-X_{i_2j_{\ell }, t_2})^2/2\sigma _d^2)$. Before calculating the Gram matrices, all covariate values are normalized to have a zero mean and a unit variance. For all $d$-dimensional Gaussian kernels, we use the same values, $\lambda _k=\lambda _d$ and $\sigma _d$. We primarily use $\lambda _k=0$ for linear kernels except for cases in which we compare models with different numbers of linear kernels. See the Results section for concrete designs of kernels. Since $\{ \alpha _k\} _{k\in \mathfrak{K}}$ has large dimensions, we use incomplete Cholesky decomposition \cite{Bach2003} to approximate $N\times N$ Gram matrix with $N\times N^{\prime }$ matrix $L_k$ as $G_k\approx L_kL_k^{\prime }$ for $N^{\prime }\ll N$. We use $0.001$ for the value of tolerance parameter in this approximation. We then have $f=\sum _kL_ku_k$ with $u_k=L_k^{\prime }\alpha _k$ and $\frac{\lambda _k}{2} \alpha _k^{\prime }G_k\alpha _k=\frac{\lambda _k}{2} \| u_k\| _2^2$. We solve the above optimization problem, by applying limited-memory BFGS algorithm \cite{Liu1989} with backtracking and line search stopped by the Armijo condition. The optimization was stopped, if the $\ell _2$-norm of the gradient gets smaller than $\epsilon _{\mathrm{stop}}=1.0\times 10^{-2}$.

For ML estimation, one can approximate the BME by using the Laplace approximation. Assigning the Hessian of the negative log-likelihood with the regularization term (\ref{min_problem}) to $\widetilde{H}$ and regarding the regularization term as the log-likelihood of a Gaussian process prior \cite{Rasmussen2006}, we calculate BME (see, e.g., Chapter 3 of Ref.\cite{Bishop2006}) using Gaussian integrals as follows:
\begin{eqnarray}
\ln \mathrm{BME}(\mathcal{M}|D)\approx -\sum _{i\in \mathcal{I}}\sum _{t\in \mathfrak{T} _i}U\left (D_{i,t}; \widehat{\theta}, \sum _{k\in \mathfrak{K}}L_k\widehat{u_k}\right )+\sum _{k\in \mathfrak{K}}\frac{\ln \lambda _k}{2}\mathrm{dim} (u_k)-\frac{1}{2}\ln |\widetilde{H}|,
\end{eqnarray} 
with the solution of Eq.(\ref{min_problem}), $\widehat{\theta } and \widehat{f}$. In the above approximation, the space of functions perpendicular to the range of $G_k$ does not contribute to $U$ (the representer theorem \cite{Berlinet2011}). Therefore, the contributions of the prior and posterior cancel out. Similarly, the space perpendicular to the range of $L_k$, but not to that of $G_k$, makes negligible contributions to $U$ relative to their variations in the prior process. In our implementation, linear functions of treatment variables, covariates and bias parameters were treated as one-dimensional functions associated with linear kernels and therefore represented as $L_ku_k$. We often use $\lambda _k=0$ for linear kernels, namely, a non-informative prior. In this case, $\ln \lambda _k$ in the BME is removed. This is justified when we compare two models with the same set of linear kernels. As we compare models with different sets of linear kernels, we need to tune $\lambda _k>0$ for the linear kernels.  

Next, we describe how the additional nuisance parameters are calculated. The Hessians $\widehat{H}_{\theta f}$ and $\widehat{H}_{ff}$ are simply obtained as the block components of the numerical Hessian with respect to $\{ u_k\} $ and $b$ (as well as $\kappa $ and $\beta $ for the latent variable model), because these coordinates are orthogonal. For the validation and held out datasets in the cross-fitting procedure, we calculated the trained solution $L_k\widehat{\alpha }_k=\widehat{u}_k$ with the aid of the pseudoinverse of $L_k$, and the decomposition $\overline{L}_k\overline{L}_k^{\prime }$ for the Gram matrix for both training and validation (held out) datasets provide new orthogonal coordinates $\overline{u} _k=\overline{L} _k\widehat{\alpha }_k$. Each of the Hessian of the negative log-likelihood of the training and validation datasets is calculated in terms of this $\overline{u} _k$ and $\mathrm{CVErr} _{H}$ is calculated according to the formula in Eq.(\ref{CVErrH}). 

For the determination of $\widehat{g}_k$, we perform optimization by replacing $U$ by  
\begin{eqnarray}
U_{g_k}(D_{i,t}; g_k)=A_{ik,t}\ln (1+e^{-g_k(X_{i,t})})+(1-\sum _{\ell }A_{i\ell ,t})\ln (1+e^{g_k(X_{i,t})})
\end{eqnarray}
in Eq.(\ref{min_problem}). The cross-validation error $\mathrm{CVErr} _{g_k}$ is also calculated via $\overline{u} _k$ according to Eq.(\ref{CVErrg}).

In the simulation with latent-variable $Z$, we apply the EM algorithm for optimization (see, e.g., Chapter 9 of Ref.\cite{Bishop2006}). Since the latent variable model has indefiniteness with respect to permutation of the range of $Z$ as do many other models with discrete latent variables, we start optimization from values close to $\theta ^*, f^*, \beta ^*$ and $\kappa ^*$. An alternately repeating sequence of updating $r(Z_i)$ with 
\begin{eqnarray}
r_i(Z)=\frac{\exp \left ( -\sum _{t\in \mathfrak{T} _i}U(D_{i,t}, Z; \theta, f, \kappa ,\beta )\right )}{\sum _{\widetilde{Z}\in \{ 0,1\} }\exp \left ( -\sum _{t\in \mathfrak{T} _i}U(D_{i,t}, \widetilde{Z}; \theta, f, \kappa ,\beta )\right )}
\end{eqnarray}
and optimizing $\theta, f, \kappa$ and $\beta $ 
\begin{eqnarray}
\min _{\theta , f, \kappa ,\beta }\sum _{i\in \mathcal{I}}\sum _{Z\in \{ 0,1\} }\sum _{t\in \mathfrak{T} _i}r_i(Z)U(D_{i,t}, Z; \theta, f,\kappa ,\beta )+\sum _{k\in \mathfrak{K}}\frac{\lambda _k}{2} \alpha _k^{\prime }G_k\alpha _k, 
\end{eqnarray} 
maximizes the marginal log-likelihood in Eq.(\ref{log_marginal_likelihood}).
In the above, the stepwise negative log-likelihood is given by 
\begin{eqnarray}
U(D_{i,t}, Z; \theta , f,\kappa ,\beta )&=&-I_{[t, t+\Delta t)}(T_i)(\theta ^{\prime }A_{it}+f(X_{it})+\kappa Z)+\exp (\theta ^{\prime }A_{it}+f(X_{it})+\kappa Z) \nonumber \\
&&+\delta _{t0}\ln \left (1+\exp \left (-(2Z-1)(\sum _j\beta _{j}X_{j0}+\beta _0)\right )\right ),
\end{eqnarray}
where Kronecker delta $\delta _{t0}$ has been used.
The procedure of estimating nuisance parameters is essentially the same as for the model without a latent variable.

For the construction of score functions with $\widehat{H}_{\theta f}$ and $\widehat{H}_{ff}$, we compute the components of $\partial _f\ell |_{\widehat{\theta }, \widehat{f}}$ proportional to $\{ e^{\theta _k}\} _k$ and the one independent of $\theta $ separately and then apply $\widehat{H} _{\theta f}(\widehat{H} _{ff} +\zeta _n)^{-1}$ to all of these components. In the case with a latent variable, we compute the components of $\partial _f\ell |_{\widehat{\theta }, \widehat{f}}$ proportional to $r_i(Z)\{ e^{\theta _k}\} $ and $r_i(Z)$ separately, and then apply $\widehat{H} _{\theta f}(\widehat{H} _{ff} +\zeta _n)^{-1}$. For the case without a latent variable, thus obtained linear equations of $\{ e^{\theta _k}\} $ can be algebraically solved. For the case with a latent variable, further numerically finding the root of score equations of the following form is necessary:
\begin{eqnarray}
\frac{1}{n}\sum _{i\in \mathcal{I}}\sum _{Z\in \{ 0, 1\} }r_i[Z](\theta _1, \theta _2)(b_{ik1}(Z)e^{\theta _1}+b_{ik2}(Z)e^{\theta _2}+b_{ik3}(Z))=0,\ \ (k\in \mathcal{K} ).
\end{eqnarray}
Solving this set of highly nonlinear and nonconvex score equations is not difficult, but we take the approach of linearizing the above equations around the initial estimate $\widehat{\theta }$ and solving the resultant linear equation. This is justifiable because the orthogonalization procedure is not designed to adjust higher-order terms around $\widehat{\theta }$ in the first place.

\section{Proof of Proposition 1} \label{appendix_prop1}
\begin{proof}
We have,
\begin{eqnarray}
&\lim _{\delta t\rightarrow +0}&\frac{ P(I_{(t,t+\delta t]}(T^a)=1|T^a, T^{a^{\prime }}>t ,A_{(-\infty , t)}=a_{(-\infty , t)}, X_{(-\infty , t+\delta t]}=x_{(-\infty , t+\delta t]})}{P(I_{(t,t+\delta t]}(T^{a^{\prime }})=1|T^a,T^{a^{\prime }}>t, A_{(-\infty , t)}=a^{\prime }_{(-\infty , t)}, X_{(-\infty , t+\delta t]}=x_{(-\infty , t+\delta t]})}  \nonumber \\
=&&\hspace{-2.0cm}\lim _{\delta t\rightarrow +0} \frac{ P(I_{(t,t+\delta t]}(T^a)=1|T^a, T^{a^{\prime }}>t ,A_{(-\infty , t+\delta t]}=a_{(-\infty , t+\delta t]}, X_{(-\infty , t+\delta t]}=x_{(-\infty , t+\delta t]})}{P(I_{(t,t+\delta t]}(T^{a^{\prime }})=1|T^a,T^{a^{\prime }}>t, A_{(-\infty , t+\delta t]}=a^{\prime }_{(-\infty , t+\delta t]}, X_{(-\infty , t+\delta t]}=x_{(-\infty , t+\delta t]})} \nonumber \\
=&&\hspace{-2.0cm}\lim _{\delta t\rightarrow +0} \frac{ P(I_{(t,t+\delta t]}(T)=1|T>t ,A_{(-\infty , t+\delta t]}=a_{(-\infty , t+\delta t]}, X_{(-\infty , t+\delta t]}=x_{(-\infty , t+\delta t]})}{P(I_{(t,t+\delta t]}(T)=1|T>t, A_{(-\infty , t+\delta t]}=a^{\prime }_{(-\infty , t+\delta t]}, X_{(-\infty , t+\delta t]}=x_{(-\infty , t+\delta t]})} \nonumber \\
=&&\hspace{-2.0cm}\lim _{\delta t\rightarrow +0}\frac{P(I_{(t,t+\delta t]}(T)=1|T>t, A_{t}=a_t, X_t=x_t)}{P(I_{(t,t+\delta t]}(T)=1|T>t, A_{t}=a_t^{\prime }, X_t=x_t)}. \label{causal_hazard_contrast}
\end{eqnarray}
The first equality is due to no unmeasured confounder (Assumption 2) thanks to which the specification of $A_{[t, t+\delta t]}$ with any values in the conditioning does not change the conditional probabilities in the numerator and denominator. With this additional conditioning, consistency (Assumption 1) ensures the equality between actual and counterfactual variables, which asserts the second equality. The last equality is obtained by first applying the regularity of hazard (Assumption 4) and then the independence of past values of treatment variables and covariates (Assumption 5).

Identifying the last line with the ML estimator is straightforward as follows. Using Assumptions 4 and 5 and Eq.(\ref{infinitesimal_hazard_likelihood}), the expected log-likelihood is represented with marginal probability measures on $\mathcal{X} \times \mathcal{A} \times \mathcal{T}$ as
\begin{eqnarray}
&&\mathrm{E} [Q_h(T|X,A)] \nonumber \\
 &=&\int _{0}^{\infty }\int _{0}^{C}\int _{\mathcal{X} \times \mathcal{A}}\lim _{\delta t\rightarrow +0}\frac{1}{\delta t}\left \{ P(I_{(t,t+\delta t]}(T)=1|T>t, A_t, X_t)(\theta ^\prime A_t +f(X_t))  \right.\nonumber \\
 && \left. -\exp (\theta ^\prime A_t+f(X_t))\delta t\right \} dP(X_t, A_t|T>t)P(T>t)dtdP(C)+const  \nonumber \\
&=&\sum _{k\in \mathcal{K}} \int _{0}^{\infty }\int _{0}^{C}\int _{\mathcal{X}}\lim _{\delta t\rightarrow +0}\frac{1}{\delta t}\left \{ P(I_{(t,t+\delta t]}(T)=1|T>t, A_{k,t}=1, X_t)(\theta _k+f(X_t)) \right. \nonumber \\
 && \left. -\exp (\theta _k+f(X_t))\delta t\right \} dP(X_t|T>t, A_{k,t}=1)P(T>t, A_{k,t}=1)dtdP(C) \nonumber \\
 && +\int _{0}^{\infty }\int _{0}^{C}\int _{\mathcal{X}} \lim _{\delta t\rightarrow +0}\frac{1}{\delta t}\left \{ P(I_{(t,t+\delta t]}(T)=1|T>t, A_t=0, X_t)f(X_t) \right. \nonumber \\
 && \left. -\exp (f(X_t))\delta t\right \} dP(X_t|T>t, A_{t}=0)P(T>t, A_{t}=0)dtdP(C) + const, 
\end{eqnarray}
where the second equality is obtained by decomposing the integration domain as $\mathcal{X} \times \{ \cup _{k\in \mathcal{K}}\{ A\ |\ A_k=1\} \cup \{ A\ |\ A=0\} \}$. 
Here, terms independent of $\theta$ and $f$ have been regarded as constants, and the integration with respect to the marginal probability measures for the data-generating process has been denoted by $dP$. Differentiating this representation gives 
\begin{eqnarray}
\hspace{-2cm}\frac{\partial }{\partial \theta _k} \mathrm{E} [Q_h(T|X,A)]&=&\int _{0}^{\infty }\int _{0}^{C}\int _{\mathcal{X} \times \mathcal{A}} \left \{ \lim _{\delta t\rightarrow +0}\frac{1}{\delta t}P(I_{(t,t+\delta t]}(T)=1|T>t, A_{k,t}=1, X_t)\right. \nonumber \\
&&\hspace{-2cm}\left. -\exp (\theta _k+f(X_t))\right \} dP(X_t|T>t, A_{k,t}=1)P(T>t, A_{k,t}=1)dtdP(C), 
\end{eqnarray}
\begin{eqnarray}
&&\hspace{-2cm}\lim _{\epsilon \rightarrow +0}\frac{1}{\epsilon } \mathrm{E} [Q_{h_{\theta, f+\epsilon \psi }}(T|X,A)-Q_{h_{\theta , f}}(T|X,A)] \nonumber \\
&=&\int _{0}^{\infty }\int _{0}^{C}\int _{\mathcal{X} \times \mathcal{A}} \left \{\lim _{\delta t\rightarrow +0} \frac{1}{\delta t}P(I_{(t,t+\delta t]}(T)=1|T>t, A_t, X_t)\psi (X_t) \right. \nonumber \\
&&\left. -\psi (X_t)\exp (\theta ^\prime A_t+f(X_t))\right \} dP(X_t, A_t|T>t)P(T>t)dtdP(C).  
\end{eqnarray}
With Assumptions 6 and 7, it is seen that these derivatives vanish if and only if 
\begin{eqnarray}
\theta _k&=&\ln \lim _{\delta t\rightarrow +0}\frac{P(I_{(t,t+\delta t]}(T)=1|T>t, A_{k,t}=1, X_t)}{P(I_{(t,t+\delta t]}(T)=1|T>t, A_t=0, X_t)}, \nonumber \\
f(X_t)&=&\ln \lim _{\delta t\rightarrow +0}\frac{1}{\delta t}P(I_{(t,t+\delta t]}(T)=1|T>t, A_t=0, X_t),
\end{eqnarray}
hold almost everywhere in $\mathcal{X} \times [0, \infty )$ with respect to $dP(X_t, A_t|T>t)P(T>t)P(C>t)dt$, and the concavity of the expected log-likelihood ensures that this solution is the maximizer.
\end{proof}

\section{Proof of Proposition 2} \label{appendix_prop2}
\begin{proof}
We refer readers to the next section for a proof of the identification of the gradients and Hessian with elements of $\mathcal{H} _k$ and an operator from $\mathcal{H} _k$ into $\mathcal{H} _k$. Then, taking the Gateaux derivative of the score with respect to $f\in \mathcal{N} _n$, we have
\begin{eqnarray}
\partial _f\phi (D; \theta ^*,(f^*,H^*))[f-f^*]&=&(H_{\theta f}^*-H_{\theta f}^*((H_{ff}^*+\zeta _n)^{-1}H_{ff}^*)(f-f^*) \nonumber \\
&=&\zeta _nH_{\theta f}^*(H_{ff}^*+\zeta _n)^{-1} (f-f^*). 
\end{eqnarray}
Using $H_{\theta _kf}^*=H_{ff}^*\rho _k$, we see that the norm of this vector is given by
\begin{eqnarray}
\zeta _n\left \{ \sum _{k\in \mathcal{K}}((H_{ff}^*+\zeta _n)^{-1}H_{ff}^*\rho _k , f-f^*)_{\mathcal{H} _k}^2\right \} ^{1/2}\sim O(n^{-\alpha -\beta}),
\end{eqnarray}
and hence the proposition is asserted for $\alpha +\beta > \frac{1}{2}$. Here we have used the fact $\| (H_{ff}^*+\zeta _n)^{-1}H_{ff}^*\| \leq 1$. 
\end{proof}

\section{Gradient functional and Hessian operator} \label{appendix_gradient_hessian}
In this section, we provide the gradient functional and the Hessian operator of the samplewise negative log-likelihood: 
\begin{eqnarray}
\ell (D; \theta ,f)=-I_{[0,C]}(T)(\theta ^{\prime }A_{T}+f(X_{T}))+\int _{0}^{T\land C} \exp\left (\theta ^{\prime }A_{t}+f(X_{t})\right ) dt. \label{negative_loglikelihood}
\end{eqnarray}

Calculating the Gateaux derivative $\frac{1}{\epsilon} (\ell (D; \theta , f+\epsilon \psi )-\ell (D; \theta , f))$ for an arbitrary element $\psi $ of the RKHS, we expect that the gradient functional is given as 
\begin{eqnarray}
\psi \in \mathcal{H} _k\ \mapsto \ -I_{[0,C]}(T)\psi (X_T)+\int _0^{T\land C}\psi (X_t)\exp \left (\theta ^{\prime }A_t+f(X_t)\right )dt\ \in \mathbf{R}.
\end{eqnarray}
This mapping is actually given as the following inner product:
\begin{eqnarray}
\psi \in \mathcal{H} _k\ \mapsto \ \left (-I_{[0,C]}(T)k(X_T, \cdot )+\int _0^{T\land C}\exp \left (\theta ^{\prime }A_t+f(X_t)\right )k(X_t,\cdot)dt, \psi \right )_{\mathcal{H}_k}\ \in \mathbf{R},
\end{eqnarray}
and satisfies the definition in the Frechet sense stated in Proposition 2.
Note that the integration of $k(X,\cdot)$ over a bounded positive measure in the left argument of the inner-product belongs to $\mathcal{H}_k$ \cite{Fukumizu2013}. Similarly, the differentiation of the above functional is given by 
\begin{eqnarray}
\psi \in \mathcal{H}_k\ \mapsto \ \int _0^{T\land C}\exp \left (\theta ^{\prime }A_t+f(X_t)\right )k(X_t,\cdot) \left (k(X_t, \cdot ), \psi \right )_{\mathcal{H}_k}dt,\ \in \mathcal{H} _k.
\end{eqnarray}
Therefore, the Hessian operator is represented as an integration of $k(X, \cdot)\otimes k(X, \cdot)$ over a bounded positive measure in a manner similar to the autocovariance opeartor \cite{Fukumizu2013}. 
Next, let us consider the case with a latent variable. In this case, we represent the samplewise negative marginal log-likelihood as 
\begin{eqnarray}
\ell (D; \theta ,f)=\sum _{Z\in \mathcal{Z}} r(Z)(\ell (D|Z; \theta, f) +\ln r(Z)), 
\end{eqnarray}
with 
\begin{eqnarray}
\ell (D|Z; \theta , f)&=&-I_{[0,C]}(T)(\theta ^{\prime }A_{T}+f_1(X_{T}, Z))\nonumber \\
&&+\int _{0}^{T\land C} \exp\left (\theta ^{\prime }A_{t}+f_1(X_{t},Z) \right ) dt-\ln Q(Z|f_2(X_0)), 
\end{eqnarray}
and 
\begin{eqnarray}
r(Z)=\frac{\exp (-\ell (D|Z; \theta , f))}{\sum _{\widetilde{Z}\in \mathcal{Z}}\exp (-\ell (D|\widetilde{Z}; \theta , f))},
\end{eqnarray}
where we have redefined $f$ as $(f_1, f_2)^{\prime }$. In the simulation study, we use $f_1(X_t,Z)=f(X_t)+\kappa Z+b$ and $f_2(X_0)=\sum _{j}\beta _jX_{j0}+\beta _0$. Direct computation shows that the gradient of this marginal negative log-likelihood is given by
\begin{eqnarray}
\partial _f\ell (D; \theta ,f)=E_{Z\sim r}[\partial _f\ell (D|Z; \theta ,f)]
\end{eqnarray}
and the Hessian is given by
\begin{eqnarray}
\partial _f\partial _f\ell (D; \theta ,f)&=&E_{Z\sim r}[\partial _f\partial _f\ell (D|Z; \theta ,f)]+E_{Z\sim r}[\partial _f\ell (D|Z; \theta ,f)]\otimes E_{Z\sim r}[\partial _f\ell (D|Z; \theta, f)] \nonumber \\
&&-E_{Z\sim r}[\partial _f\ell (D|Z; \theta, f)\otimes \partial _f\ell (D|Z; \theta, f)].
\end{eqnarray}

\section{The validity of Assumption 9} \label{appendix_assumption9}
As we have seen in the previous section, $\widehat{H_{\theta f}}$ and $\widehat{H_{ff}}$ are represented as the sum of terms of the forms, $\widehat{E} [\int \exp (\widehat{\theta }^{\prime }A_t+\widehat{f}(X_t)) A_t\otimes k(X_t, \cdot )dt]$ and $\widehat{E} [\int \exp (\widehat{\theta }^{\prime }A_t+\widehat{f}(X_t)) k(X_t, \cdot )\otimes k(X_t, \cdot )dt]$, respectively. For latent variable model, the empirical estimators also involve terms such as $\widehat{E} [(I_{0,C}(T) k(X_T, \cdot)-\int \exp (\widehat{\theta }^{\prime }A_t+\widehat{f}(X_t)) k(X_t, \cdot )dt)\otimes (I_{0,C}(T) k(X_T, \cdot)-\int \exp (\widehat{\theta }^{\prime }A_t+\widehat{f}(X_t)) k(X_t, \cdot )dt)]$. The $O_p(n^{-1/2})$ norm-convergence of empirically averaged operators of the same form but without the integration of a stochastic process to their large-sample limit are known (see Sec 9.1 of Ref.\cite{Berlinet2011}; and \cite{Fukumizu2007} for a stronger convergence in Hilbert-Schmidt norm). We believe that a similar convergence result can be obtained, but we leave this as a conjecture in order to avoid the argument about the stochastic process $(A_{it}, X_{it})_t$, which is not the focus of the current study. From the representation of the Hessian operator, it is also justified that 
\begin{eqnarray}
\left \|E[\partial _{(\theta ,f)}\partial _{(\theta ,f)}\ell | _{\widehat{\theta }, \widehat{f}}]-H^* \right \| \sim O_p(n^{-\beta }), 
\end{eqnarray}
because $f^*$ is a bounded function and we have
\begin{eqnarray}
\sup _{X\in \mathcal{X}}|f(X)-f^*(X)|=\sup _{X\in \mathcal{X}}(f-f^*, k(X, \cdot ))_{\mathcal{H} _k}\leq \sup _{X\in \mathcal{X}}\| f-f^*\| _{\mathcal{H} _k}k(X,X)^{1/2}.
\end{eqnarray}

From the representation of the Hessian, it is also reasonably justified that $H_{f\theta _k}^*=H_{ff}^*\rho _k$ holds. Let us define the following marginal probability measures, for $^\forall E\in \mathfrak{M} _{\mathcal{X}}$,
\begin{eqnarray}
\nu _0(E)=\int _0^{\infty }\int _0^C\int _E\exp (\theta ^{\prime }A_t+f(X_t))dP(X_t, A_t|T>t)P(T>t)dtdP(C),
\end{eqnarray}
\begin{eqnarray}
\nu _k(E)=\int _0^{\infty }\int _0^C\int _EA_k\exp (\theta ^{\prime }A_t+f(X_t))dP(X_t, A_t|T>t)P(T>t)dtdP(C).
\end{eqnarray}
Suppose that $\nu _k$ is absolutely continuous with respect to $\nu _0$ and the Radom-Nykodym derivative $\frac{d\nu _k}{d\nu _0}=\rho _k$ belongs to $\mathcal{H} _k$. Then, we can see $H_{f\theta _k}^*=H_{ff}^*\rho _k$. This scenario is plausible enough. It should be noted that the samplewise Hessian $\partial _{\theta _k}\partial _f\ell $ itself is very unlikely to lie in the range of $H_{ff}^*$. Comparing the above result with Proposition 3, one can see that, roughly speaking, the two methods based on Proposition 2 and 3 are performing density-ratio estimation in two different ways, namely, least-square regression and logistic regression (cf: see \cite{Kanamori2010} for multiple ways of density-ratio estimation based on the kernel method).

For $H_{\theta f}^*$ and $H_{ff} ^*$ of the latent-variable model, we cannot represent $\rho _k$ as a density ratio, but it is still plausible enough that $H_{\theta _kf}^*$ and the functions in the range of $H_{ff}^*$ are the integral of $k(X,\cdot)$ over some smooth density on $\mathcal{X}$ in commonly studied settings and $H_{\theta _k}^*=H_{ff}^*\rho _k$ holds.

\section{Proof of Proposition 3} \label{appendix_prop3}
\begin{proof}
Consider the following hazard model that includes the heterogeneous treatment effect (i.e., covariate-dependent treatment effect): 
\begin{eqnarray}
\widetilde{h} _{\theta ,f,f_{\mathrm{het}}}(t|A_t,X_t)=\exp (\theta ^\prime A_t+f(X_t)+f_{\mathrm{het}}(X_t)A_{k,t}). 
\end{eqnarray}
Because we assumed the homogeneity of the treatment effect (Assumption 6), we have, for arbitrary $\psi $, 
\begin{eqnarray}
&\lim _{\epsilon \rightarrow +0}&\frac{1}{\epsilon }\mathrm{E} \left [\ln Q_{\widetilde{h} _{\theta ^*, f^*, \epsilon \psi }}(T|A,X)-\ln Q_{\widetilde{h} _{\theta ^*, f^*,0}}(T|A,X)\right ] \nonumber \\
&&=\mathrm{E} \left [I_{[0,C]}(T)A_{k,T}\psi (X_T)-\int _0^{T\land C} A_{k,t}\psi (X_t)\exp (\theta _{k}^*A_{k,t}+f^*(X_t))dt\right ] \nonumber \\
&&=0. \label{HE_derivative}
\end{eqnarray}
Note that, Assumption 8 assures that the argument of the Gateaux derivative $\psi $ does not need to belong to $\mathcal{M}$. 
Let the first line on the right-hand side of Eq.(\ref{ML_scores}) be $\phi _{k,1}(D; \theta, (f,g_k))$ and the second line be $\phi _{k,2}(D; \theta, (f,g_k))$. Note that $\mathrm{E} [\phi _{k,1}]$ with $\theta _k=\theta _k^*$ and $f=f^*$ coincides with the above equation with $\psi (X_t)=e^{-\theta _k^*}(1+e^{-g_k(X_t)})$. Similarly, for arbitrary $\psi $, we have 
\begin{eqnarray}
&\lim _{\epsilon \rightarrow +0}&\frac{1}{\epsilon }\mathrm{E}\left [\ln Q_{h_{\theta ^*, f^*+\epsilon \psi }}(T|A,X)-\ln Q_{h_{\theta ^*, f^*}}(T|A,X)\right ] \nonumber \\
&&=\mathrm{E} \left [I_{[0,C]}(T)\psi (X_T)-\int _0^{T\land C} \psi (X_t)\exp (\theta ^{*\prime }A_{t}+f^*(X_t))dt\right ] \nonumber \\
&&=0. \label{uniform_derivative}
\end{eqnarray}
Subtracting the sum of Eq.(\ref{HE_derivative}) for all $k$ from Eq.(\ref{uniform_derivative}) for $\psi (X_t)=1+e^{g_k(X_t)}$, we obtain $\mathrm{E} [\phi _{k,2}(D; \theta ^*, (f^*, g_k))]=0$. Hence, $E[\phi _k(D; \theta ^*, (f^*, g_k))]=0$ has been proven. 

Next, we examine the Gateaux derivatives of the expected scores with respect to the nuisance parameters $(f,g_k)$.
First, the derivative of $\mathrm{E} [\phi _{k}]$ with respect to $g_k$ at $\theta _k=\theta _k^*$ and $f=f^*$ is
\begin{eqnarray}
&\lim _{\epsilon \rightarrow +0} &\frac{1}{\epsilon } \mathrm{E} \left [\phi _k(D; \theta ^*, (f^*, g_k+\epsilon \psi ))-\phi _k(D; \theta ^*, (f^*, g_k))\right ] \nonumber \\
&=&\hspace{-0.8cm}\mathrm{E} \left [-I_{[0,C]}(T)A_{k,T}\psi (X_T)e^{-\theta _k^*-g_k(X_T)}+\int _{0}^{T\land C}A_{k,t}\psi (X_t)e^{f^*(X_{t})-g_k(X_{t})}dt \right ] \nonumber \\
&&\hspace{-0.8cm}+\mathrm{E} \left [\int _{0}^{T\land C}(1-\sum _{\ell \in \mathcal{K}}A_{\ell ,t})\psi (X_t)e^{f^*(X_{t})+g_k(X_t)}dt-(1-\sum _{\ell \in \mathcal{K}}A_{\ell ,T})\psi (X_T)e^{g_k(X_T)}dt\right ] \nonumber \\
&=&0.
\end{eqnarray}
The combination of Eqs.(\ref{HE_derivative}) and (\ref{uniform_derivative}) again proves the last equality. 
The derivative with respect to $f$ at $\theta =\theta ^*$, $f=f^*$ and $g_k=g_k^*$ is given by 
\begin{eqnarray}
&\lim _{\epsilon \rightarrow +0} &\left. \frac{1}{\epsilon } \mathrm{E} \left [\phi _k(D; \theta ^*, (f^*+\epsilon \psi , g_k^*)-\phi _k(D; \theta ^*, (f^*, g_k^*))\right ] \right |_{f=f^*}\nonumber \\
=&&\hspace{-1.6cm}\mathrm{E} \left [\int _{0}^{T\land C}(1-\sum _{\ell \in \mathcal{K}}A_{\ell ,t})\psi (X_t)e^{f^*(X_{t})}(1+e^{g_k^*(X_t)})-A_{k,t}\psi (X_t)e^{f^*(X_{t})}(1+e^{-g_k^*(X_{t})})dt\right ] \nonumber \\
&=&\hspace{-0.7cm}0. \label{score_f_derivative}
\end{eqnarray}
Eq.(\ref{definition_gk}) shows that the expected values of the two integrands in the second line cancel out, which proves the last equality.
\end{proof}

\section{Consideration on the score regularity and the quality of estimation of nuisance parameters required for DML} \label{appendix_score_regularity}
Chernozhukov et al. \cite{Chernozhukov2018} provided additional conditions for score regularity and convergence of nuisance parameters so that debiasing works (Assumptions 3.3 and 3.4 of their paper). Assumption 3.3 imposes a straightforward regularity condition for non-degeneracy concerning $\partial _\theta \mathrm{E} [\phi (D; \theta, \eta ^*)|_{\theta ^*}]$, which we can reasonably assume for ML estimation with identifiable models. Assumption 3.4 additionally imposes the following nontrivial conditions, together with conventionally used conditions on the covering number and non-degeneracy of the model. 

\noindent { \bf Proposition 4 [Assumption 3.4(c): Chernozhukov et al. 2018)] } {\it Assume that $\Theta $ is bounded, and the suitably defined norm of all derivatives of the expected log-likelihood up to the third order with respect to $\theta $ and $f$ are bounded in the product set of $\Theta $ and $\mathcal{N}_{\overline{n}}$ (for some $\overline{n}\in \mathbf{N}$). Concretely, define a tensor
\begin{eqnarray}
\partial _f\partial _f\partial _f\ell \ : (\psi _1,\psi _2) \in \mathcal{H} _k\times \mathcal{H} _k\ \mapsto \ \psi _3\in \mathcal{H} _k,
\end{eqnarray}
in the same manner as for the gradients and Hessian, and assume $\| \partial _f\partial _f\partial _f\ell |_{\theta \in \Theta ,f\in \mathcal{N} _{\overline{n}}}(\psi _1,\psi _2)\| _{\mathcal{H} _k}\leq C\| \psi _1\| _{\mathcal{H} _k}\| \psi _2\| _{\mathcal{H} _k}$ for some positive constant $C$. Also make the same sort of assumptions for the other derivatives as well.

Then, with three sequences of positive constants converging to zero, $\{ \delta _n\} _n$, $\{ \Delta _n\} _n$ and $\{ \tau _n\} _n$, the following three conditions are satisfied: 
\begin{eqnarray}
&&\sup _{\eta \in \mathcal{T} _n, \theta \in \Theta } \| E[\phi (D; \theta , \eta )-\phi (D; \theta , \eta ^*)] \| \leq \delta _n\tau _n, \nonumber \\
r_n&=&\sup _{\eta \in \mathcal{T} _n, \| \theta -\theta ^*\| \leq \tau _n}  E[\| \phi (D; \theta , \eta )-\phi (D; \theta^*, \eta ^*)\| ^2]^{1/2}\ and\ r_n\ln ^{1/2}(1/r_n)\leq \delta _n,\nonumber  \\
&&\sup _{r\in (0, 1), \eta \in \mathcal{T} _n, \| \theta -\theta ^*\|\leq \tau _n} \| \partial _r^2E[\phi (D; \theta ^*+r(\theta -\theta ^*), \eta ^*+r(\eta -\eta ^*))]\| \leq \delta _nn^{-1/2}. \label{score_regularity_conv}
\end{eqnarray}
}
\begin{proof}
The quantity in the first line is bounded as 
\begin{eqnarray}
&&\| E[\phi (D; \theta , \eta )-\phi (D; \theta , \eta ^*)] \| \nonumber \\
&=&\| E[\partial _{\theta }\ell |_{\theta,f} -H_{\theta f}(H_{ff}+\zeta _n)^{-1}\partial _{f}\ell |_{\theta ,f} -\partial _{\theta }\ell|_{\theta ,f^*}+H_{\theta f}^*(H_{ff}^*+\zeta _n)^{-1}\partial _{f}\ell |_{\theta ,f^*}]\| \nonumber \\
&\leq &\| E[\partial _{\theta }\ell |_{\theta,f}-\partial _{\theta }\ell |_{\theta ,f^*}]\| +\| E[H_{\theta f}(H_{ff}+\zeta _n)^{-1}(\partial _{f}\ell |_{\theta,f}-\partial _{f}\ell|_{\theta ,f^*})]\| \nonumber \\
&+& \| E[(H_{\theta f}-H_{\theta _f}^*)(H_{ff}^*+\zeta _n)^{-1}\partial _f\ell | _{\theta ,f^*}]\| \nonumber \\
&+& \| E[H_{\theta f}(H_{ff}+\zeta _n)^{-1}(H_{ff}-H_{ff}^*)(H_{ff}^*+\zeta _n)^{-1}\partial _f\ell | _{\theta ,f^*}]\| \label{first_cond}
\end{eqnarray}
For the evaluation of each term, we first use 
\begin{eqnarray}
H_{\theta f}(H_{ff}+\zeta _n)^{-1}&=&\lim _{m\rightarrow \infty }H_{\theta f, m}(H_{ff, m}+\zeta _n)^{-1} \nonumber \\
&=&\lim _{m\rightarrow \infty }H_{\theta \theta , m}^{1/2}W_{\theta f,m}H_{ff, m}^{1/2}(H_{ff, m}+\zeta _n)^{-1} \nonumber \\
&\sim & O(n^{\frac{\alpha }{2}}).
\end{eqnarray}
Here, we have represented the compact positive-semidefinite operator $H$ as the limit of a sequence of positive-definite finite-rank operators (therefore represented as matrices), and then we have used the decomposition $H_{\theta f, m}=H_{\theta \theta , m}^{1/2}W_{\theta f,m}H_{ff, m}^{1/2}$ with $\| W_{\theta f, m}\| \leq 1$. Then, we can evaluate the terms of the upper bound in Eq.(\ref{first_cond}) as $O(n^{-\beta })$, $O(n^{\frac{\alpha }{2}-\beta })$, $O(n^{\alpha-\beta })$ and $O(n^{\frac{3\alpha }{2}-\beta })$. From these conditions, one can see that $\beta >\frac{3}{2} \alpha $ is required.

The norm within the expectation of the second line of Eqs.(\ref{score_regularity_conv}) is similarly bounded as
\begin{eqnarray}
&&E\|\phi (D; \theta , \eta )-\phi (D; \theta ^*, \eta ^*)\| ^2\nonumber \\
&=&E\| \partial _{\theta }\ell |_{\theta,f} -H_{\theta f}(H_{ff}+\zeta _n)^{-1}\partial _{f}\ell |_{\theta ,f} -\partial _{\theta }\ell|_{\theta ^*,f^*}+H_{\theta f}^*(H_{ff}^*+\zeta _n)^{-1}\partial _{f}\ell |_{\theta ^*,f^*}\| ^2\nonumber \\
&\leq &E\| \partial _{\theta }\ell |_{\theta,f}-\partial _{\theta }\ell |_{\theta ^*,f^*}\| ^2+E\| H_{\theta f}(H_{ff}+\zeta _n)^{-1}(\partial _{f}\ell |_{\theta,f}-\partial _{f}\ell|_{\theta ^*,f^*})\| ^2\nonumber \\
&+& E\| (H_{\theta f}-H_{\theta _f}^*)(H_{ff}^*+\zeta _n)^{-1}\partial _f\ell | _{\theta ^*,f^*}\| ^2\nonumber \\
&+& E\| H_{\theta f}(H_{ff}+\zeta _n)^{-1}(H_{ff}-H_{ff}^*)(H_{ff}^*+\zeta _n)^{-1}\partial _f\ell | _{\theta ^*,f^*}\| ^2,
\end{eqnarray}
and similarly evaluated as $O(\tau _n^2)$, $O(n^{\alpha }\tau _n^2)$, $O(n^{2(-\beta +\alpha )}$ and $O(n^{3\alpha -2\beta })$. Together with the condition obtained above, we have $n^{\frac{3}{2}\alpha -\beta }<\tau _n<n^{-\frac{1}{2}\alpha }$ which requires $\beta >2\alpha $. 

Finally the upperbound for the expected value of the twice Gateaux derivative is written down with $\delta \theta =(\theta -\theta ^*)$ and $\delta f=(f-f^*)$ etc., together with slight notational abuse for brevity:
\begin{eqnarray}
&&\| \partial _r^2E[\phi (D; \theta ^*+r(\theta -\theta ^*), \eta ^*+r(\eta -\eta ^*))]\| \nonumber \\
&\leq& \|E[\partial _{\theta \theta \theta} \ell \delta \theta \delta \theta ]\| _2+ 2\|E[\partial _{\theta \theta f}\ell \delta \theta \delta f ]\| _2+\|E[\partial _{\theta ff}\ell \delta f\delta f]\| _2\nonumber \\
&+&2\| E[\delta H_{\theta f}(H_{ff}+\zeta _n)^{-1}(\partial _{f\theta }\ell \delta \theta +\partial _{ff}\ell \delta f)]\|\nonumber\\
&+&2\| E[H_{\theta f}(H_{ff}+\zeta _n)^{-1}\delta H_{ff}(H_{ff}+\zeta _n)^{-1}\delta H_{ff}(H_{ff}+\zeta _n)^{-1}\partial _f\ell ]\| _2\nonumber \\
&+&2\| E[H_{\theta f}(H_{ff}+\zeta _n)^{-1}\delta H_{ff}(H_{ff}+\zeta _n)^{-1}(\partial _{f\theta }\ell \delta \theta +\partial _{ff}\ell \delta f)]\| _2\nonumber \\
&+&\| E[H_{\theta f}(H_{ff}+\zeta _n)^{-1}(\partial _{f\theta\theta}\ell \delta \theta \delta \theta +2\partial _{f\theta f}\ell \delta \theta \delta f+\partial _{fff}\ell \delta f\delta f)]\| _2.
\end{eqnarray}
Evaluating each term in the same manner as for the previous conditions, we conclude $\tau _nn^{\frac{3}{2}\alpha -\beta }, n^{\frac{3}{2}\alpha -2\beta }, \tau _n^2n^{\frac{\alpha }{2}}\sim o(n^{-\frac{1}{2}})$. Inserting the lower bound of $\tau _n$ with respect to $\alpha $ and $\beta $ due to the second condition, one can see that $\alpha $ should satisfy $\frac{1}{2}-\beta <\alpha <\frac{4}{7}\beta -\frac{1}{7}, \frac{1}{2}\beta $, which gives a condition $\beta >\frac{63}{154}$. The existence of a set of values for $\alpha $ and $\beta $ satisfying the above conditions proves the assertion. Here, note that terms with convergence slower than a power of $n$ in the assertion does not take effect as long as we deal with strict inequalities of exponents of $n$.
\end{proof}

\noindent {\bf Remark 7.}  For the orthogonal score presented in Proposition 3, the relationship between the above conditions and the convergence of $f$ and $g_k$ can be easily seen. Thus, we omit writing it down to avoid redundancy.

\section{Additional consideration on causal interpretation of hazard ratios} \label{appendix_causal_interpretation}

\noindent {\bf Remark 8-1.} Suppose that Assumption 2 holds for arbitrary $\epsilon >0$. Then, for arbitrary $t>t^{\prime }$, we obtain
\begin{eqnarray}
\lim _{\delta t\rightarrow +0}\frac{1}{\delta t}P(I_{(t, t+\delta t]}(T^a)=1|T^a>t, X_{(-\infty , t+\delta t]}, A_{(-\infty ,t^{\prime })}=a_{(-\infty ,t^{\prime })})\nonumber \\
=\lim _{\delta t\rightarrow +0}\frac{1}{\delta t}P(I_{(t, t+\delta t]}(T)=1|T>t, A_t=a_t, X_t),
\end{eqnarray}
with the aid of Assumptions 1--5 in the same manner as we have done for Eqs.(\ref{ignorability_conseq}) and (\ref{causal_hazard_contrast}). If we can assume $A_{(-\infty , t^{\prime })}=a_{(-\infty , t^{\prime })}=0$ with probability one at some $t^{\prime }$, the above equation implies that the counterfactual survival function is given by
\begin{eqnarray}
S^a(t)=\mathrm{E} \left [\exp \left (-\int _0^t\exp (\theta ^{*\prime }a_s+f^*(X_s))ds\right )\right ].
\end{eqnarray}
Although measures of causal effects can be calculated from this relation, doing so requires an unbiased estimation of $f^*$, which is possible, only if a simple model can be used to estimate $f$ without modeling error. 

\noindent {\bf Remark 8-2.} As discussed in Martinussen et al. (2022) \cite{Martinussen2022}, of more interest is the quantity 
\begin{eqnarray}
\lim _{\delta t\rightarrow +0}\ln \frac{ P(I_{(t,t+\delta t]}(T^{a})=1|T^{a},T^{a^{\prime }}>t ,X^a_{(-\infty , t+\delta t]}=X^{a^{\prime }}_{(-\infty , t+\delta t]})}{P(I_{(t,t+\delta t]}(T^{a^{\prime }})=1|T^{a},T^{a^{\prime }}>t ,X^a_{(-\infty , t+\delta t]}=X^{a^{\prime }}_{(-\infty , t+\delta t]})}, \label{strong_causal_effect}
\end{eqnarray}
for two arbitrary counterfactual treatment schedules that branch before $t$. A stronger untestable assumption is required for this to be related to the hazard ratios. For example,
\begin{eqnarray} 
T \mathop{\perp\!\!\!\!\perp}  T^a, X^a&|&T>t, \{ A_s\} _{s\geq t}, \{ X_s\} _{s\geq t} \nonumber \\
T^a \mathop{\perp\!\!\!\!\perp}  A, T, X&|&T^a>t, \{ X_s^a\} _{s\geq t},  \nonumber \\
T^a \mathop{\perp\!\!\!\!\perp}  T^{a^{\prime }}, X^{a^{\prime}}&|&T^{a}>t, \{ X_s^{a}\} _{s\geq t}, \label{strong_assumption}
\end{eqnarray}
for any $t$ and any counterfactual treatment schedules $a,a^{\prime }$. This condition is satisfied if no unmeasured factors affect the outcome, as described by the causal graph in Figure 1(A). If unmeasured factors affect the outcome, as described by the causal graph in Figure 1(B), the above conditions are not satisfied. Although the assumption of time-homogeneity (Assumption 7) is strong and does not allow temporal change in the heterogeneity of risk to be left unmodeled, it does not completely exclude the existence of unmeasured factors affecting outcome. For example, suppose that the survival of subjects for an arbitrary counterfactual treatment schedule $a$ is described by 
\begin{eqnarray}
P(T_i^a>t)=\exp \left (-\int _0^t\exp (\theta ^{\prime *} a_{s}+f^*(X_{i,s}))(1+W_{i,s})ds\right ), \ \ X_{i,s}=X_{i,s}^a,\ \ W_{i,s}=W_{i,s}^a,
\end{eqnarray}
where $W_{i,t}=\sum _j\delta (t-t_{ij})$ is instantaneous random fluctuations in the conditional hazard described by Dirac delta functions centered at time points determined by a homogeneous Poisson point process $\{ t_{ij}\} _j$. The subject-wise random effect of $W_i$ shared between $T, T^a$ and $T^{a^{\prime }}$ then violates Eq.(\ref{strong_assumption}), whereas $W_i$ at different time points are independent and do not influence the time-homogeneity. 

\begin{figure}[!ht]
\centering
\includegraphics[width=100mm]{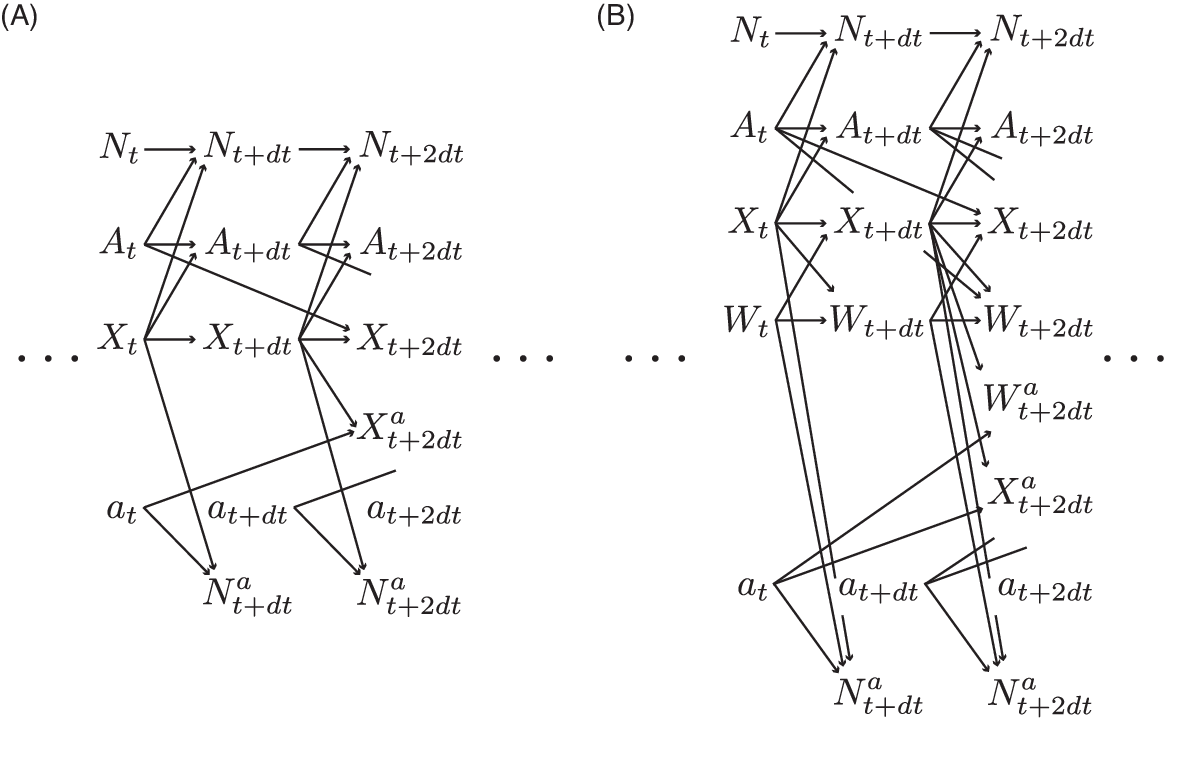}
\caption{Two causal graphs describing the causal relationships among the counting process for outcome event, $N_t=I_{(-\infty , t]}(T)$, treatment variable, $A_t$, covariate, $X_t$ and unmeasured factors, $W_t$ which bifurcate into an actual process and a counterfactual process at time $t$. Remark 3 discusses how these causal relationships affect the interpretation of the estimation results in the proposed framework.}
\label{figA}
\end{figure}

\bibliography{ref.bib}

\begin{thebibliography}{59}%
\makeatletter
\providecommand \@ifxundefined [1]{%
 \@ifx{#1\undefined}
}%
\providecommand \@ifnum [1]{%
 \ifnum #1\expandafter \@firstoftwo
 \else \expandafter \@secondoftwo
 \fi
}%
\providecommand \@ifx [1]{%
 \ifx #1\expandafter \@firstoftwo
 \else \expandafter \@secondoftwo
 \fi
}%
\providecommand \natexlab [1]{#1}%
\providecommand \enquote  [1]{``#1''}%
\providecommand \bibnamefont  [1]{#1}%
\providecommand \bibfnamefont [1]{#1}%
\providecommand \citenamefont [1]{#1}%
\providecommand \href@noop [0]{\@secondoftwo}%
\providecommand \href [0]{\begingroup \@sanitize@url \@href}%
\providecommand \@href[1]{\@@startlink{#1}\@@href}%
\providecommand \@@href[1]{\endgroup#1\@@endlink}%
\providecommand \@sanitize@url [0]{\catcode `\\12\catcode `\$12\catcode
  `\&12\catcode `\#12\catcode `\^12\catcode `\_12\catcode `\%12\relax}%
\providecommand \@@startlink[1]{}%
\providecommand \@@endlink[0]{}%
\providecommand \url  [0]{\begingroup\@sanitize@url \@url }%
\providecommand \@url [1]{\endgroup\@href {#1}{\urlprefix }}%
\providecommand \urlprefix  [0]{URL }%
\providecommand \Eprint [0]{\href }%
\providecommand \doibase [0]{http://dx.doi.org/}%
\providecommand \selectlanguage [0]{\@gobble}%
\providecommand \bibinfo  [0]{\@secondoftwo}%
\providecommand \bibfield  [0]{\@secondoftwo}%
\providecommand \translation [1]{[#1]}%
\providecommand \BibitemOpen [0]{}%
\providecommand \bibitemStop [0]{}%
\providecommand \bibitemNoStop [0]{.\EOS\space}%
\providecommand \EOS [0]{\spacefactor3000\relax}%
\providecommand \BibitemShut  [1]{\csname bibitem#1\endcsname}%
\let\auto@bib@innerbib\@empty
\bibitem [{\citenamefont {Lin}\ \emph {et~al.}(2020)\citenamefont {Lin},
  \citenamefont {Lin}, \citenamefont {Roychoudhury}, \citenamefont {Anderson},
  \citenamefont {Hu}, \citenamefont {Huang}, \citenamefont {Leon},
  \citenamefont {Liao}, \citenamefont {Liu}, \citenamefont {Luo}, \citenamefont
  {Mukhopadhyay}, \citenamefont {Qin}, \citenamefont {Tatsuoka}, \citenamefont
  {Wang}, \citenamefont {Wang}, \citenamefont {Zhu}, \citenamefont {Chen},\
  and\ \citenamefont {and}}]{Lin2020}%
  \BibitemOpen
  \bibfield  {author} {\bibinfo {author} {\bibfnamefont {Ray~S.}\ \bibnamefont
  {Lin}}, \bibinfo {author} {\bibfnamefont {Ji}~\bibnamefont {Lin}}, \bibinfo
  {author} {\bibfnamefont {Satrajit}\ \bibnamefont {Roychoudhury}}, \bibinfo
  {author} {\bibfnamefont {Keaven~M.}\ \bibnamefont {Anderson}}, \bibinfo
  {author} {\bibfnamefont {Tianle}\ \bibnamefont {Hu}}, \bibinfo {author}
  {\bibfnamefont {Bo}~\bibnamefont {Huang}}, \bibinfo {author} {\bibfnamefont
  {Larry~F}\ \bibnamefont {Leon}}, \bibinfo {author} {\bibfnamefont
  {Jason~J.Z.}\ \bibnamefont {Liao}}, \bibinfo {author} {\bibfnamefont {Rong}\
  \bibnamefont {Liu}}, \bibinfo {author} {\bibfnamefont {Xiaodong}\
  \bibnamefont {Luo}}, \bibinfo {author} {\bibfnamefont {Pralay}\ \bibnamefont
  {Mukhopadhyay}}, \bibinfo {author} {\bibfnamefont {Rui}\ \bibnamefont {Qin}},
  \bibinfo {author} {\bibfnamefont {Kay}\ \bibnamefont {Tatsuoka}}, \bibinfo
  {author} {\bibfnamefont {Xuejing}\ \bibnamefont {Wang}}, \bibinfo {author}
  {\bibfnamefont {Yang}\ \bibnamefont {Wang}}, \bibinfo {author} {\bibfnamefont
  {Jian}\ \bibnamefont {Zhu}}, \bibinfo {author} {\bibfnamefont {Tai-Tsang}\
  \bibnamefont {Chen}}, \ and\ \bibinfo {author} {\bibfnamefont {Renee~Iacona}\
  \bibnamefont {and}},\ }\bibfield  {title} {\enquote {\bibinfo {title}
  {Alternative analysis methods for time to event endpoints under
  nonproportional hazards: A comparative analysis},}\ }\href {\doibase
  10.1080/19466315.2019.1697738} {\bibfield  {journal} {\bibinfo  {journal}
  {Statistics in Biopharmaceutical Research}\ }\textbf {\bibinfo {volume}
  {12}},\ \bibinfo {pages} {187--198} (\bibinfo {year} {2020})},\ \Eprint
  {http://arxiv.org/abs/https://doi.org/10.1080/19466315.2019.1697738}
  {https://doi.org/10.1080/19466315.2019.1697738} \BibitemShut {NoStop}%
\bibitem [{\citenamefont {Bartlett}\ \emph {et~al.}(2020)\citenamefont
  {Bartlett}, \citenamefont {Morris}, \citenamefont {Stensrud}, \citenamefont
  {Daniel}, \citenamefont {Vansteelandt},\ and\ \citenamefont
  {Burman}}]{Bartlett2020}%
  \BibitemOpen
  \bibfield  {author} {\bibinfo {author} {\bibfnamefont {Jonathan~W.}\
  \bibnamefont {Bartlett}}, \bibinfo {author} {\bibfnamefont {Tim~P.}\
  \bibnamefont {Morris}}, \bibinfo {author} {\bibfnamefont {Mats~J.}\
  \bibnamefont {Stensrud}}, \bibinfo {author} {\bibfnamefont {Rhian~M.}\
  \bibnamefont {Daniel}}, \bibinfo {author} {\bibfnamefont {Stijn~K.}\
  \bibnamefont {Vansteelandt}}, \ and\ \bibinfo {author} {\bibfnamefont
  {Carl-Fredrik}\ \bibnamefont {Burman}},\ }\bibfield  {title} {\enquote
  {\bibinfo {title} {The hazards of period specific and weighted hazard
  ratios},}\ }\href {\doibase 10.1080/19466315.2020.1755722} {\bibfield
  {journal} {\bibinfo  {journal} {Statistics in Biopharmaceutical Research}\
  }\textbf {\bibinfo {volume} {12}},\ \bibinfo {pages} {518--519} (\bibinfo
  {year} {2020})},\ \Eprint
  {http://arxiv.org/abs/https://doi.org/10.1080/19466315.2020.1755722}
  {https://doi.org/10.1080/19466315.2020.1755722} \BibitemShut {NoStop}%
\bibitem [{\citenamefont {Hern{\'a}n}(2010)}]{Hernan2010}%
  \BibitemOpen
  \bibfield  {author} {\bibinfo {author} {\bibfnamefont {Miguel~A.}\
  \bibnamefont {Hern{\'a}n}},\ }\bibfield  {title} {\enquote {\bibinfo {title}
  {The hazards of hazard ratios},}\ }\href
  {https://journals.lww.com/epidem/fulltext/2010/01000/the_hazards_of_hazard_ratios.4.aspx}
  {\bibfield  {journal} {\bibinfo  {journal} {Epidemiology}\ }\textbf {\bibinfo
  {volume} {21}} (\bibinfo {year} {2010})}\BibitemShut {NoStop}%
\bibitem [{\citenamefont {Aalen}\ \emph {et~al.}(2015)\citenamefont {Aalen},
  \citenamefont {Cook},\ and\ \citenamefont {R{\o}ysland}}]{Aalen2015}%
  \BibitemOpen
  \bibfield  {author} {\bibinfo {author} {\bibfnamefont {Odd~O.}\ \bibnamefont
  {Aalen}}, \bibinfo {author} {\bibfnamefont {Richard~J.}\ \bibnamefont
  {Cook}}, \ and\ \bibinfo {author} {\bibfnamefont {Kjetil}\ \bibnamefont
  {R{\o}ysland}},\ }\bibfield  {title} {\enquote {\bibinfo {title} {Does cox
  analysis of a randomized survival study yield a causal treatment effect?}}\
  }\href {\doibase 10.1007/s10985-015-9335-y} {\bibfield  {journal} {\bibinfo
  {journal} {Lifetime Data Analysis}\ }\textbf {\bibinfo {volume} {21}},\
  \bibinfo {pages} {579--593} (\bibinfo {year} {2015})}\BibitemShut {NoStop}%
\bibitem [{\citenamefont {Martinussen}\ \emph {et~al.}(2020)\citenamefont
  {Martinussen}, \citenamefont {Vansteelandt},\ and\ \citenamefont
  {Andersen}}]{Martinussen2020}%
  \BibitemOpen
  \bibfield  {author} {\bibinfo {author} {\bibfnamefont {Torben}\ \bibnamefont
  {Martinussen}}, \bibinfo {author} {\bibfnamefont {Stijn}\ \bibnamefont
  {Vansteelandt}}, \ and\ \bibinfo {author} {\bibfnamefont {Per~Kragh}\
  \bibnamefont {Andersen}},\ }\bibfield  {title} {\enquote {\bibinfo {title}
  {Subtleties in the interpretation of hazard contrasts},}\ }\href {\doibase
  10.1007/s10985-020-09501-5} {\bibfield  {journal} {\bibinfo  {journal}
  {Lifetime Data Analysis}\ }\textbf {\bibinfo {volume} {26}},\ \bibinfo
  {pages} {833--855} (\bibinfo {year} {2020})}\BibitemShut {NoStop}%
\bibitem [{\citenamefont {Martinussen}(2022)}]{Martinussen2022}%
  \BibitemOpen
  \bibfield  {author} {\bibinfo {author} {\bibfnamefont {Torben}\ \bibnamefont
  {Martinussen}},\ }\bibfield  {title} {\enquote {\bibinfo {title} {Causality
  and the cox regression model},}\ }\href {\doibase
  https://doi.org/10.1146/annurev-statistics-040320-114441} {\bibfield
  {journal} {\bibinfo  {journal} {Annual Review of Statistics and Its
  Application}\ }\textbf {\bibinfo {volume} {9}},\ \bibinfo {pages} {249--259}
  (\bibinfo {year} {2022})}\BibitemShut {NoStop}%
\bibitem [{\citenamefont {Prentice}\ and\ \citenamefont
  {Aragaki}(2022)}]{Prentice2022}%
  \BibitemOpen
  \bibfield  {author} {\bibinfo {author} {\bibfnamefont {Ross~L}\ \bibnamefont
  {Prentice}}\ and\ \bibinfo {author} {\bibfnamefont {Aaron~K}\ \bibnamefont
  {Aragaki}},\ }\bibfield  {title} {\enquote {\bibinfo {title}
  {Intention-to-treat comparisons in randomized trials},}\ }\href@noop {}
  {\bibfield  {journal} {\bibinfo  {journal} {Statistical Science}\ }\textbf
  {\bibinfo {volume} {37}},\ \bibinfo {pages} {380--393} (\bibinfo {year}
  {2022})}\BibitemShut {NoStop}%
\bibitem [{\citenamefont {Ying}\ and\ \citenamefont {Xu}(2023)}]{Ying2023}%
  \BibitemOpen
  \bibfield  {author} {\bibinfo {author} {\bibfnamefont {Andrew}\ \bibnamefont
  {Ying}}\ and\ \bibinfo {author} {\bibfnamefont {Ronghui}\ \bibnamefont
  {Xu}},\ }\href {https://arxiv.org/abs/2307.11971} {\enquote {\bibinfo {title}
  {On defense of the hazard ratio},}\ } (\bibinfo {year} {2023}),\ \Eprint
  {http://arxiv.org/abs/2307.11971} {arXiv:2307.11971 [math.ST]} \BibitemShut
  {NoStop}%
\bibitem [{\citenamefont {Fay}\ and\ \citenamefont {Li}(2024)}]{Fay2024}%
  \BibitemOpen
  \bibfield  {author} {\bibinfo {author} {\bibfnamefont {Michael~P}\
  \bibnamefont {Fay}}\ and\ \bibinfo {author} {\bibfnamefont {Fan}\
  \bibnamefont {Li}},\ }\bibfield  {title} {\enquote {\bibinfo {title} {Causal
  interpretation of the hazard ratio in randomized clinical trials},}\ }\href
  {\doibase 10.1177/17407745241243308} {\bibfield  {journal} {\bibinfo
  {journal} {Clinical Trials}\ }\textbf {\bibinfo {volume} {21}},\ \bibinfo
  {pages} {623--635} (\bibinfo {year} {2024})},\ \bibinfo {note} {pMID:
  38679930},\ \Eprint
  {http://arxiv.org/abs/https://doi.org/10.1177/17407745241243308}
  {https://doi.org/10.1177/17407745241243308} \BibitemShut {NoStop}%
\bibitem [{\citenamefont {Rufibach}(2019)}]{Rufibach2019}%
  \BibitemOpen
  \bibfield  {author} {\bibinfo {author} {\bibfnamefont {Kaspar}\ \bibnamefont
  {Rufibach}},\ }\bibfield  {title} {\enquote {\bibinfo {title} {Treatment
  effect quantification for time-to-event endpoints–estimands, analysis
  strategies, and beyond},}\ }\href {\doibase https://doi.org/10.1002/pst.1917}
  {\bibfield  {journal} {\bibinfo  {journal} {Pharmaceutical Statistics}\
  }\textbf {\bibinfo {volume} {18}},\ \bibinfo {pages} {145--165} (\bibinfo
  {year} {2019})},\ \Eprint
  {http://arxiv.org/abs/https://onlinelibrary.wiley.com/doi/pdf/10.1002/pst.1917}
  {https://onlinelibrary.wiley.com/doi/pdf/10.1002/pst.1917} \BibitemShut
  {NoStop}%
\bibitem [{\citenamefont {Kloecker}\ \emph {et~al.}(2020)\citenamefont
  {Kloecker}, \citenamefont {Davies}, \citenamefont {Khunti},\ and\
  \citenamefont {Zaccardi}}]{Kloecker2020}%
  \BibitemOpen
  \bibfield  {author} {\bibinfo {author} {\bibfnamefont {David~E.}\
  \bibnamefont {Kloecker}}, \bibinfo {author} {\bibfnamefont {Melanie~J.}\
  \bibnamefont {Davies}}, \bibinfo {author} {\bibfnamefont {Kamlesh.}\
  \bibnamefont {Khunti}}, \ and\ \bibinfo {author} {\bibfnamefont {Francesco}\
  \bibnamefont {Zaccardi}},\ }\bibfield  {title} {\enquote {\bibinfo {title}
  {Uses and limitations of the restricted mean survival time: Illustrative
  examples from cardiovascular outcomes and mortality trials in type 2
  diabetes},}\ }\href {\doibase 10.7326/M19-3286} {\bibfield  {journal}
  {\bibinfo  {journal} {Annals of Internal Medicine}\ }\textbf {\bibinfo
  {volume} {172}},\ \bibinfo {pages} {541--552} (\bibinfo {year} {2020})},\
  \bibinfo {note} {pMID: 32203984},\ \Eprint
  {http://arxiv.org/abs/https://doi.org/10.7326/M19-3286}
  {https://doi.org/10.7326/M19-3286} \BibitemShut {NoStop}%
\bibitem [{\citenamefont {Snapinn}\ \emph {et~al.}(2023)\citenamefont
  {Snapinn}, \citenamefont {Jiang},\ and\ \citenamefont {Ke}}]{Snapinn2023}%
  \BibitemOpen
  \bibfield  {author} {\bibinfo {author} {\bibfnamefont {Steven}\ \bibnamefont
  {Snapinn}}, \bibinfo {author} {\bibfnamefont {Qi}~\bibnamefont {Jiang}}, \
  and\ \bibinfo {author} {\bibfnamefont {Chunlei}\ \bibnamefont {Ke}},\
  }\bibfield  {title} {\enquote {\bibinfo {title} {Treatment effect measures
  under nonproportional hazards},}\ }\href {\doibase
  https://doi.org/10.1002/pst.2267} {\bibfield  {journal} {\bibinfo  {journal}
  {Pharmaceutical Statistics}\ }\textbf {\bibinfo {volume} {22}},\ \bibinfo
  {pages} {181--193} (\bibinfo {year} {2023})},\ \Eprint
  {http://arxiv.org/abs/https://onlinelibrary.wiley.com/doi/pdf/10.1002/pst.2267}
  {https://onlinelibrary.wiley.com/doi/pdf/10.1002/pst.2267} \BibitemShut
  {NoStop}%
\bibitem [{\citenamefont {Cui}\ \emph {et~al.}(2023)\citenamefont {Cui},
  \citenamefont {Kosorok}, \citenamefont {Sverdrup}, \citenamefont {Wager},\
  and\ \citenamefont {Zhu}}]{Cui2023}%
  \BibitemOpen
  \bibfield  {author} {\bibinfo {author} {\bibfnamefont {Yifan}\ \bibnamefont
  {Cui}}, \bibinfo {author} {\bibfnamefont {Michael~R}\ \bibnamefont
  {Kosorok}}, \bibinfo {author} {\bibfnamefont {Erik}\ \bibnamefont
  {Sverdrup}}, \bibinfo {author} {\bibfnamefont {Stefan}\ \bibnamefont
  {Wager}}, \ and\ \bibinfo {author} {\bibfnamefont {Ruoqing}\ \bibnamefont
  {Zhu}},\ }\bibfield  {title} {\enquote {\bibinfo {title} {Estimating
  heterogeneous treatment effects with right-censored data via causal survival
  forests},}\ }\href {\doibase 10.1093/jrsssb/qkac001} {\bibfield  {journal}
  {\bibinfo  {journal} {Journal of the Royal Statistical Society Series B:
  Statistical Methodology}\ }\textbf {\bibinfo {volume} {85}},\ \bibinfo
  {pages} {179--211} (\bibinfo {year} {2023})},\ \Eprint
  {http://arxiv.org/abs/https://academic.oup.com/jrsssb/article-pdf/85/2/179/50204704/qkac001.pdf}
  {https://academic.oup.com/jrsssb/article-pdf/85/2/179/50204704/qkac001.pdf}
  \BibitemShut {NoStop}%
\bibitem [{\citenamefont {Xu}\ \emph {et~al.}(2024)\citenamefont {Xu},
  \citenamefont {Cobzaru}, \citenamefont {Finkelstein}, \citenamefont {Welsch},
  \citenamefont {Ng},\ and\ \citenamefont {Shahn}}]{Xu2024}%
  \BibitemOpen
  \bibfield  {author} {\bibinfo {author} {\bibfnamefont {Shenbo}\ \bibnamefont
  {Xu}}, \bibinfo {author} {\bibfnamefont {Raluca}\ \bibnamefont {Cobzaru}},
  \bibinfo {author} {\bibfnamefont {Stan~N.}\ \bibnamefont {Finkelstein}},
  \bibinfo {author} {\bibfnamefont {Roy~E.}\ \bibnamefont {Welsch}}, \bibinfo
  {author} {\bibfnamefont {Kenney}\ \bibnamefont {Ng}}, \ and\ \bibinfo
  {author} {\bibfnamefont {Zach}\ \bibnamefont {Shahn}},\ }\href
  {https://arxiv.org/abs/2401.11263} {\enquote {\bibinfo {title} {Estimating
  heterogeneous treatment effects on survival outcomes using counterfactual
  censoring unbiased transformations},}\ } (\bibinfo {year} {2024}),\ \Eprint
  {http://arxiv.org/abs/2401.11263} {arXiv:2401.11263 [stat.ME]} \BibitemShut
  {NoStop}%
\bibitem [{\citenamefont {Frauen}\ \emph {et~al.}(2025)\citenamefont {Frauen},
  \citenamefont {Schröder}, \citenamefont {Hess},\ and\ \citenamefont
  {Feuerriegel}}]{Frauen2025}%
  \BibitemOpen
  \bibfield  {author} {\bibinfo {author} {\bibfnamefont {Dennis}\ \bibnamefont
  {Frauen}}, \bibinfo {author} {\bibfnamefont {Maresa}\ \bibnamefont
  {Schröder}}, \bibinfo {author} {\bibfnamefont {Konstantin}\ \bibnamefont
  {Hess}}, \ and\ \bibinfo {author} {\bibfnamefont {Stefan}\ \bibnamefont
  {Feuerriegel}},\ }\href {https://arxiv.org/abs/2505.13072} {\enquote
  {\bibinfo {title} {Orthogonal survival learners for estimating heterogeneous
  treatment effects from time-to-event data},}\ } (\bibinfo {year} {2025}),\
  \Eprint {http://arxiv.org/abs/2505.13072} {arXiv:2505.13072 [cs.LG]}
  \BibitemShut {NoStop}%
\bibitem [{\citenamefont {Leviton}\ and\ \citenamefont
  {Loddenkemper}(2023)}]{Leviton2023}%
  \BibitemOpen
  \bibfield  {author} {\bibinfo {author} {\bibfnamefont {Alan}\ \bibnamefont
  {Leviton}}\ and\ \bibinfo {author} {\bibfnamefont {Tobias}\ \bibnamefont
  {Loddenkemper}},\ }\bibfield  {title} {\enquote {\bibinfo {title} {Design,
  implementation, and inferential issues associated with clinical trials that
  rely on data in electronic medical records: a narrative review},}\ }\href
  {\doibase 10.1186/s12874-023-02102-4} {\bibfield  {journal} {\bibinfo
  {journal} {BMC Medical Research Methodology}\ }\textbf {\bibinfo {volume}
  {23}},\ \bibinfo {pages} {271} (\bibinfo {year} {2023})}\BibitemShut
  {NoStop}%
\bibitem [{\citenamefont {Hernán}\ \emph {et~al.}(2001)\citenamefont
  {Hernán}, \citenamefont {Brumback},\ and\ \citenamefont
  {Robins}}]{Hernan2001}%
  \BibitemOpen
  \bibfield  {author} {\bibinfo {author} {\bibfnamefont {Miguel~A}\
  \bibnamefont {Hernán}}, \bibinfo {author} {\bibfnamefont {Babette}\
  \bibnamefont {Brumback}}, \ and\ \bibinfo {author} {\bibfnamefont {James~M}\
  \bibnamefont {Robins}},\ }\bibfield  {title} {\enquote {\bibinfo {title}
  {Marginal structural models to estimate the joint causal effect of
  nonrandomized treatments},}\ }\href {\doibase 10.1198/016214501753168154}
  {\bibfield  {journal} {\bibinfo  {journal} {Journal of the American
  Statistical Association}\ }\textbf {\bibinfo {volume} {96}},\ \bibinfo
  {pages} {440--448} (\bibinfo {year} {2001})},\ \Eprint
  {http://arxiv.org/abs/https://doi.org/10.1198/016214501753168154}
  {https://doi.org/10.1198/016214501753168154} \BibitemShut {NoStop}%
\bibitem [{\citenamefont {Van~der Laan}\ and\ \citenamefont
  {Rose}(2018)}]{vanderLaan2018}%
  \BibitemOpen
  \bibfield  {author} {\bibinfo {author} {\bibfnamefont {Mark~J}\ \bibnamefont
  {Van~der Laan}}\ and\ \bibinfo {author} {\bibfnamefont {Sherri}\ \bibnamefont
  {Rose}},\ }\href@noop {} {\emph {\bibinfo {title} {Targeted learning in data
  science}}}\ (\bibinfo  {publisher} {Springer},\ \bibinfo {year}
  {2018})\BibitemShut {NoStop}%
\bibitem [{\citenamefont {Chernozhukov}\ \emph {et~al.}(2018)\citenamefont
  {Chernozhukov}, \citenamefont {Chetverikov}, \citenamefont {Demirer},
  \citenamefont {Duflo}, \citenamefont {Hansen}, \citenamefont {Newey},\ and\
  \citenamefont {Robins}}]{Chernozhukov2018}%
  \BibitemOpen
  \bibfield  {author} {\bibinfo {author} {\bibfnamefont {Victor}\ \bibnamefont
  {Chernozhukov}}, \bibinfo {author} {\bibfnamefont {Denis}\ \bibnamefont
  {Chetverikov}}, \bibinfo {author} {\bibfnamefont {Mert}\ \bibnamefont
  {Demirer}}, \bibinfo {author} {\bibfnamefont {Esther}\ \bibnamefont {Duflo}},
  \bibinfo {author} {\bibfnamefont {Christian}\ \bibnamefont {Hansen}},
  \bibinfo {author} {\bibfnamefont {Whitney}\ \bibnamefont {Newey}}, \ and\
  \bibinfo {author} {\bibfnamefont {James}\ \bibnamefont {Robins}},\ }\bibfield
   {title} {\enquote {\bibinfo {title} {Double/debiased machine learning for
  treatment and structural parameters},}\ }\href {\doibase
  https://doi.org/10.1111/ectj.12097} {\bibfield  {journal} {\bibinfo
  {journal} {The Econometrics Journal}\ }\textbf {\bibinfo {volume} {21}},\
  \bibinfo {pages} {C1--C68} (\bibinfo {year} {2018})},\ \Eprint
  {http://arxiv.org/abs/https://onlinelibrary.wiley.com/doi/pdf/10.1111/ectj.12097}
  {https://onlinelibrary.wiley.com/doi/pdf/10.1111/ectj.12097} \BibitemShut
  {NoStop}%
\bibitem [{\citenamefont {Ahrens}\ \emph {et~al.}(2025)\citenamefont {Ahrens},
  \citenamefont {Chernozhukov}, \citenamefont {Hansen}, \citenamefont {Kozbur},
  \citenamefont {Schaffer},\ and\ \citenamefont {Wiemann}}]{Ahrens2025}%
  \BibitemOpen
  \bibfield  {author} {\bibinfo {author} {\bibfnamefont {Achim}\ \bibnamefont
  {Ahrens}}, \bibinfo {author} {\bibfnamefont {Victor}\ \bibnamefont
  {Chernozhukov}}, \bibinfo {author} {\bibfnamefont {Christian}\ \bibnamefont
  {Hansen}}, \bibinfo {author} {\bibfnamefont {Damian}\ \bibnamefont {Kozbur}},
  \bibinfo {author} {\bibfnamefont {Mark}\ \bibnamefont {Schaffer}}, \ and\
  \bibinfo {author} {\bibfnamefont {Thomas}\ \bibnamefont {Wiemann}},\ }\href
  {https://arxiv.org/abs/2504.08324} {\enquote {\bibinfo {title} {An
  introduction to double/debiased machine learning},}\ } (\bibinfo {year}
  {2025}),\ \Eprint {http://arxiv.org/abs/2504.08324} {arXiv:2504.08324
  [econ.EM]} \BibitemShut {NoStop}%
\bibitem [{\citenamefont {Ren}\ and\ \citenamefont {Zhou}(2011)}]{Ren2011}%
  \BibitemOpen
  \bibfield  {author} {\bibinfo {author} {\bibfnamefont {Jian-Jian}\
  \bibnamefont {Ren}}\ and\ \bibinfo {author} {\bibfnamefont {Mai}\
  \bibnamefont {Zhou}},\ }\bibfield  {title} {\enquote {\bibinfo {title} {Full
  likelihood inferences in the cox model: an empirical likelihood approach},}\
  }\href {\doibase 10.1007/s10463-010-0272-y} {\bibfield  {journal} {\bibinfo
  {journal} {Annals of the Institute of Statistical Mathematics}\ }\textbf
  {\bibinfo {volume} {63}},\ \bibinfo {pages} {1005--1018} (\bibinfo {year}
  {2011})}\BibitemShut {NoStop}%
\bibitem [{\citenamefont {Berlinet}\ and\ \citenamefont
  {Thomas-Agnan}(2011)}]{Berlinet2011}%
  \BibitemOpen
  \bibfield  {author} {\bibinfo {author} {\bibfnamefont {Alain}\ \bibnamefont
  {Berlinet}}\ and\ \bibinfo {author} {\bibfnamefont {Christine}\ \bibnamefont
  {Thomas-Agnan}},\ }\href@noop {} {\emph {\bibinfo {title} {Reproducing kernel
  Hilbert spaces in probability and statistics}}}\ (\bibinfo  {publisher}
  {Springer Science \& Business Media},\ \bibinfo {year} {2011})\BibitemShut
  {NoStop}%
\bibitem [{\citenamefont {Fukumizu}\ \emph {et~al.}(2013)\citenamefont
  {Fukumizu}, \citenamefont {Song},\ and\ \citenamefont
  {Gretton}}]{Fukumizu2013}%
  \BibitemOpen
  \bibfield  {author} {\bibinfo {author} {\bibfnamefont {Kenji}\ \bibnamefont
  {Fukumizu}}, \bibinfo {author} {\bibfnamefont {Le}~\bibnamefont {Song}}, \
  and\ \bibinfo {author} {\bibfnamefont {Arthur}\ \bibnamefont {Gretton}},\
  }\bibfield  {title} {\enquote {\bibinfo {title} {Kernel bayes' rule: Bayesian
  inference with positive definite kernels},}\ }\href
  {http://jmlr.org/papers/v14/fukumizu13a.html} {\bibfield  {journal} {\bibinfo
   {journal} {Journal of Machine Learning Research}\ }\textbf {\bibinfo
  {volume} {14}},\ \bibinfo {pages} {3753--3783} (\bibinfo {year}
  {2013})}\BibitemShut {NoStop}%
\bibitem [{\citenamefont {Yang}\ \emph {et~al.}(2014)\citenamefont {Yang},
  \citenamefont {Eaton}, \citenamefont {Lu},\ and\ \citenamefont
  {Lapane}}]{Yang2014}%
  \BibitemOpen
  \bibfield  {author} {\bibinfo {author} {\bibfnamefont {Shibing}\ \bibnamefont
  {Yang}}, \bibinfo {author} {\bibfnamefont {Charles~B.}\ \bibnamefont
  {Eaton}}, \bibinfo {author} {\bibfnamefont {Juan}\ \bibnamefont {Lu}}, \ and\
  \bibinfo {author} {\bibfnamefont {Kate~L.}\ \bibnamefont {Lapane}},\
  }\bibfield  {title} {\enquote {\bibinfo {title} {Application of marginal
  structural models in pharmacoepidemiologic studies: a systematic review},}\
  }\href {\doibase https://doi.org/10.1002/pds.3569} {\bibfield  {journal}
  {\bibinfo  {journal} {Pharmacoepidemiology and Drug Safety}\ }\textbf
  {\bibinfo {volume} {23}},\ \bibinfo {pages} {560--571} (\bibinfo {year}
  {2014})},\ \Eprint
  {http://arxiv.org/abs/https://onlinelibrary.wiley.com/doi/pdf/10.1002/pds.3569}
  {https://onlinelibrary.wiley.com/doi/pdf/10.1002/pds.3569} \BibitemShut
  {NoStop}%
\bibitem [{\citenamefont {Robins}\ \emph {et~al.}(2000)\citenamefont {Robins},
  \citenamefont {Hern{\'a}n},\ and\ \citenamefont {Brumback}}]{Robins2000}%
  \BibitemOpen
  \bibfield  {author} {\bibinfo {author} {\bibfnamefont {James~M.}\
  \bibnamefont {Robins}}, \bibinfo {author} {\bibfnamefont {Miguel~{\'A}ngel}\
  \bibnamefont {Hern{\'a}n}}, \ and\ \bibinfo {author} {\bibfnamefont
  {Babette}\ \bibnamefont {Brumback}},\ }\bibfield  {title} {\enquote {\bibinfo
  {title} {Marginal structural models and causal inference in epidemiology},}\
  }\href
  {https://journals.lww.com/epidem/fulltext/2000/09000/marginal_structural_models_and_causal_inference_in.11.aspx}
  {\bibfield  {journal} {\bibinfo  {journal} {Epidemiology}\ }\textbf {\bibinfo
  {volume} {11}} (\bibinfo {year} {2000})}\BibitemShut {NoStop}%
\bibitem [{\citenamefont {Bishop}\ and\ \citenamefont
  {Nasrabadi}(2006)}]{Bishop2006}%
  \BibitemOpen
  \bibfield  {author} {\bibinfo {author} {\bibfnamefont {Christopher~M}\
  \bibnamefont {Bishop}}\ and\ \bibinfo {author} {\bibfnamefont {Nasser~M}\
  \bibnamefont {Nasrabadi}},\ }\href@noop {} {\emph {\bibinfo {title} {Pattern
  recognition and machine learning}}},\ Vol.~\bibinfo {volume} {4}\ (\bibinfo
  {publisher} {Springer},\ \bibinfo {year} {2006})\BibitemShut {NoStop}%
\bibitem [{\citenamefont {van~der Laan}\ \emph {et~al.}(2005)\citenamefont
  {van~der Laan}, \citenamefont {Petersen},\ and\ \citenamefont
  {Joffe}}]{vanderLaan2005}%
  \BibitemOpen
  \bibfield  {author} {\bibinfo {author} {\bibfnamefont {Mark~J}\ \bibnamefont
  {van~der Laan}}, \bibinfo {author} {\bibfnamefont {Maya~L}\ \bibnamefont
  {Petersen}}, \ and\ \bibinfo {author} {\bibfnamefont {Marshall~M}\
  \bibnamefont {Joffe}},\ }\bibfield  {title} {\enquote {\bibinfo {title}
  {History-adjusted marginal structural models and statically-optimal dynamic
  treatment regimens},}\ }\href@noop {} {\bibfield  {journal} {\bibinfo
  {journal} {The International Journal of Biostatistics}\ }\textbf {\bibinfo
  {volume} {1}} (\bibinfo {year} {2005})}\BibitemShut {NoStop}%
\bibitem [{\citenamefont {Hille}\ and\ \citenamefont
  {Phillips}(1974)}]{Hille1974}%
  \BibitemOpen
  \bibfield  {author} {\bibinfo {author} {\bibfnamefont {Einar}\ \bibnamefont
  {Hille}}\ and\ \bibinfo {author} {\bibfnamefont {Ralph~S}\ \bibnamefont
  {Phillips}},\ }\bibfield  {title} {\enquote {\bibinfo {title} {Functional
  analysis and semi-groups, 3rd printing of rev. ed. of 1957},}\ }in\
  \href@noop {} {\emph {\bibinfo {booktitle} {Colloq. Publ}}},\ Vol.~\bibinfo
  {volume} {31}\ (\bibinfo {year} {1974})\BibitemShut {NoStop}%
\bibitem [{\citenamefont {Lanckriet}\ \emph {et~al.}(2004)\citenamefont
  {Lanckriet}, \citenamefont {Cristianini}, \citenamefont {Bartlett},
  \citenamefont {Ghaoui},\ and\ \citenamefont {Jordan}}]{Lanckriet2004}%
  \BibitemOpen
  \bibfield  {author} {\bibinfo {author} {\bibfnamefont {Gert~RG}\ \bibnamefont
  {Lanckriet}}, \bibinfo {author} {\bibfnamefont {Nello}\ \bibnamefont
  {Cristianini}}, \bibinfo {author} {\bibfnamefont {Peter}\ \bibnamefont
  {Bartlett}}, \bibinfo {author} {\bibfnamefont {Laurent~El}\ \bibnamefont
  {Ghaoui}}, \ and\ \bibinfo {author} {\bibfnamefont {Michael~I}\ \bibnamefont
  {Jordan}},\ }\bibfield  {title} {\enquote {\bibinfo {title} {Learning the
  kernel matrix with semidefinite programming},}\ }\href@noop {} {\bibfield
  {journal} {\bibinfo  {journal} {Journal of Machine Learning Research}\
  }\textbf {\bibinfo {volume} {5}},\ \bibinfo {pages} {27--72} (\bibinfo {year}
  {2004})}\BibitemShut {NoStop}%
\bibitem [{\citenamefont {Suzuki}\ and\ \citenamefont
  {Sugiyama}(2013)}]{Suzuki2013}%
  \BibitemOpen
  \bibfield  {author} {\bibinfo {author} {\bibfnamefont {Taiji}\ \bibnamefont
  {Suzuki}}\ and\ \bibinfo {author} {\bibfnamefont {Masashi}\ \bibnamefont
  {Sugiyama}},\ }\bibfield  {title} {\enquote {\bibinfo {title} {{Fast learning
  rate of multiple kernel learning: Trade-off between sparsity and
  smoothness}},}\ }\href {\doibase 10.1214/13-AOS1095} {\bibfield  {journal}
  {\bibinfo  {journal} {The Annals of Statistics}\ }\textbf {\bibinfo {volume}
  {41}},\ \bibinfo {pages} {1381 -- 1405} (\bibinfo {year} {2013})}\BibitemShut
  {NoStop}%
\bibitem [{\citenamefont {Aronszajn}(1950)}]{Aronszajn1950}%
  \BibitemOpen
  \bibfield  {author} {\bibinfo {author} {\bibfnamefont {Nachman}\ \bibnamefont
  {Aronszajn}},\ }\bibfield  {title} {\enquote {\bibinfo {title} {Theory of
  reproducing kernels},}\ }\href@noop {} {\bibfield  {journal} {\bibinfo
  {journal} {Transactions of the American mathematical society}\ }\textbf
  {\bibinfo {volume} {68}},\ \bibinfo {pages} {337--404} (\bibinfo {year}
  {1950})}\BibitemShut {NoStop}%
\bibitem [{\citenamefont {Bach}(2008)}]{Bach2008}%
  \BibitemOpen
  \bibfield  {author} {\bibinfo {author} {\bibfnamefont {Francis~R}\
  \bibnamefont {Bach}},\ }\bibfield  {title} {\enquote {\bibinfo {title}
  {Consistency of the group lasso and multiple kernel learning.}}\ }\href@noop
  {} {\bibfield  {journal} {\bibinfo  {journal} {Journal of Machine Learning
  Research}\ }\textbf {\bibinfo {volume} {9}} (\bibinfo {year}
  {2008})}\BibitemShut {NoStop}%
\bibitem [{\citenamefont {Meier}\ \emph {et~al.}(2009)\citenamefont {Meier},
  \citenamefont {Van~de Geer},\ and\ \citenamefont {B{\"u}hlmann}}]{Meier2009}%
  \BibitemOpen
  \bibfield  {author} {\bibinfo {author} {\bibfnamefont {Lukas}\ \bibnamefont
  {Meier}}, \bibinfo {author} {\bibfnamefont {Sara}\ \bibnamefont {Van~de
  Geer}}, \ and\ \bibinfo {author} {\bibfnamefont {Peter}\ \bibnamefont
  {B{\"u}hlmann}},\ }\bibfield  {title} {\enquote {\bibinfo {title}
  {High-dimensional additive modeling},}\ }\href@noop {} {\bibfield  {journal}
  {\bibinfo  {journal} {The Annals of Statistics}\ }\textbf {\bibinfo {volume}
  {37}},\ \bibinfo {pages} {3779--3821} (\bibinfo {year} {2009})}\BibitemShut
  {NoStop}%
\bibitem [{\citenamefont {Koltchinskii}\ and\ \citenamefont
  {Yuan}(2010)}]{Koltchinskii2010}%
  \BibitemOpen
  \bibfield  {author} {\bibinfo {author} {\bibfnamefont {Vladimir}\
  \bibnamefont {Koltchinskii}}\ and\ \bibinfo {author} {\bibfnamefont {Ming}\
  \bibnamefont {Yuan}},\ }\bibfield  {title} {\enquote {\bibinfo {title}
  {Sparsity in multiple kernel learning},}\ }\href@noop {} {\bibfield
  {journal} {\bibinfo  {journal} {The Annals of Statistics}\ }\textbf {\bibinfo
  {volume} {38}},\ \bibinfo {pages} {3660--3695} (\bibinfo {year}
  {2010})}\BibitemShut {NoStop}%
\bibitem [{\citenamefont {Cox}(1972)}]{Cox1972}%
  \BibitemOpen
  \bibfield  {author} {\bibinfo {author} {\bibfnamefont {D.~R.}\ \bibnamefont
  {Cox}},\ }\bibfield  {title} {\enquote {\bibinfo {title} {Regression models
  and life-tables},}\ }\href {\doibase
  https://doi.org/10.1111/j.2517-6161.1972.tb00899.x} {\bibfield  {journal}
  {\bibinfo  {journal} {Journal of the Royal Statistical Society: Series B
  (Methodological)}\ }\textbf {\bibinfo {volume} {34}},\ \bibinfo {pages}
  {187--202} (\bibinfo {year} {1972})},\ \Eprint
  {http://arxiv.org/abs/https://rss.onlinelibrary.wiley.com/doi/pdf/10.1111/j.2517-6161.1972.tb00899.x}
  {https://rss.onlinelibrary.wiley.com/doi/pdf/10.1111/j.2517-6161.1972.tb00899.x}
  \BibitemShut {NoStop}%
\bibitem [{\citenamefont {Efron}(1977)}]{Efron1977}%
  \BibitemOpen
  \bibfield  {author} {\bibinfo {author} {\bibfnamefont {Bradley}\ \bibnamefont
  {Efron}},\ }\bibfield  {title} {\enquote {\bibinfo {title} {The efficiency of
  cox's likelihood function for censored data},}\ }\href
  {http://www.jstor.org/stable/2286217} {\bibfield  {journal} {\bibinfo
  {journal} {Journal of the American Statistical Association}\ }\textbf
  {\bibinfo {volume} {72}},\ \bibinfo {pages} {557--565} (\bibinfo {year}
  {1977})}\BibitemShut {NoStop}%
\bibitem [{\citenamefont {Oakes}(1977)}]{Oakes1977}%
  \BibitemOpen
  \bibfield  {author} {\bibinfo {author} {\bibfnamefont {David}\ \bibnamefont
  {Oakes}},\ }\bibfield  {title} {\enquote {\bibinfo {title} {The asymptotic
  information in censored survival data},}\ }\href
  {http://www.jstor.org/stable/2345319} {\bibfield  {journal} {\bibinfo
  {journal} {Biometrika}\ }\textbf {\bibinfo {volume} {64}},\ \bibinfo {pages}
  {441--448} (\bibinfo {year} {1977})}\BibitemShut {NoStop}%
\bibitem [{\citenamefont {Thackham}\ and\ \citenamefont
  {Ma}(2020)}]{Thackham2020}%
  \BibitemOpen
  \bibfield  {author} {\bibinfo {author} {\bibfnamefont {Mark}\ \bibnamefont
  {Thackham}}\ and\ \bibinfo {author} {\bibfnamefont {Jun}\ \bibnamefont
  {Ma}},\ }\bibfield  {title} {\enquote {\bibinfo {title} {On maximum
  likelihood estimation of the semi-parametric cox model with time-varying
  covariates},}\ }\href@noop {} {\bibfield  {journal} {\bibinfo  {journal}
  {Journal of Applied Statistics}\ }\textbf {\bibinfo {volume} {47}},\ \bibinfo
  {pages} {1511--1528} (\bibinfo {year} {2020})}\BibitemShut {NoStop}%
\bibitem [{\citenamefont {Luo}\ \emph {et~al.}(2024)\citenamefont {Luo},
  \citenamefont {Rava}, \citenamefont {Bradic},\ and\ \citenamefont
  {Xu}}]{Luo2024}%
  \BibitemOpen
  \bibfield  {author} {\bibinfo {author} {\bibfnamefont {Jiyu}\ \bibnamefont
  {Luo}}, \bibinfo {author} {\bibfnamefont {Denise}\ \bibnamefont {Rava}},
  \bibinfo {author} {\bibfnamefont {Jelena}\ \bibnamefont {Bradic}}, \ and\
  \bibinfo {author} {\bibfnamefont {Ronghui}\ \bibnamefont {Xu}},\ }\bibfield
  {title} {\enquote {\bibinfo {title} {Doubly robust estimation under a
  possibly misspecified marginal structural cox model},}\ }\href {\doibase
  10.1093/biomet/asae065} {\bibfield  {journal} {\bibinfo  {journal}
  {Biometrika}\ }\textbf {\bibinfo {volume} {112}},\ \bibinfo {pages} {asae065}
  (\bibinfo {year} {2024})},\ \Eprint
  {http://arxiv.org/abs/https://academic.oup.com/biomet/article-pdf/112/1/asae065/60791527/asae065.pdf}
  {https://academic.oup.com/biomet/article-pdf/112/1/asae065/60791527/asae065.pdf}
  \BibitemShut {NoStop}%
\bibitem [{\citenamefont {Zhang}\ \emph {et~al.}(2023)\citenamefont {Zhang},
  \citenamefont {Stringer}, \citenamefont {Brown},\ and\ \citenamefont
  {Stafford}}]{Zhang2023}%
  \BibitemOpen
  \bibfield  {author} {\bibinfo {author} {\bibfnamefont {Ziang}\ \bibnamefont
  {Zhang}}, \bibinfo {author} {\bibfnamefont {Alex}\ \bibnamefont {Stringer}},
  \bibinfo {author} {\bibfnamefont {Patrick}\ \bibnamefont {Brown}}, \ and\
  \bibinfo {author} {\bibfnamefont {Jamie}\ \bibnamefont {Stafford}},\
  }\bibfield  {title} {\enquote {\bibinfo {title} {Bayesian inference for cox
  proportional hazard models with partial likelihoods, nonlinear covariate
  effects and correlated observations},}\ }\href {\doibase
  10.1177/09622802221134172} {\bibfield  {journal} {\bibinfo  {journal}
  {Statistical Methods in Medical Research}\ }\textbf {\bibinfo {volume}
  {32}},\ \bibinfo {pages} {165--180} (\bibinfo {year} {2023})},\ \bibinfo
  {note} {pMID: 36317395},\ \Eprint
  {http://arxiv.org/abs/https://doi.org/10.1177/09622802221134172}
  {https://doi.org/10.1177/09622802221134172} \BibitemShut {NoStop}%
\bibitem [{\citenamefont {Inoue}\ \emph {et~al.}(2024)\citenamefont {Inoue},
  \citenamefont {Adomi}, \citenamefont {Efthimiou}, \citenamefont {Komura},
  \citenamefont {Omae}, \citenamefont {Onishi}, \citenamefont {Tsutsumi},
  \citenamefont {Fujii}, \citenamefont {Kondo},\ and\ \citenamefont
  {Furukawa}}]{Inoue2024}%
  \BibitemOpen
  \bibfield  {author} {\bibinfo {author} {\bibfnamefont {Kosuke}\ \bibnamefont
  {Inoue}}, \bibinfo {author} {\bibfnamefont {Motohiko}\ \bibnamefont {Adomi}},
  \bibinfo {author} {\bibfnamefont {Orestis}\ \bibnamefont {Efthimiou}},
  \bibinfo {author} {\bibfnamefont {Toshiaki}\ \bibnamefont {Komura}}, \bibinfo
  {author} {\bibfnamefont {Kenji}\ \bibnamefont {Omae}}, \bibinfo {author}
  {\bibfnamefont {Akira}\ \bibnamefont {Onishi}}, \bibinfo {author}
  {\bibfnamefont {Yusuke}\ \bibnamefont {Tsutsumi}}, \bibinfo {author}
  {\bibfnamefont {Tomoko}\ \bibnamefont {Fujii}}, \bibinfo {author}
  {\bibfnamefont {Naoki}\ \bibnamefont {Kondo}}, \ and\ \bibinfo {author}
  {\bibfnamefont {Toshi~A.}\ \bibnamefont {Furukawa}},\ }\bibfield  {title}
  {\enquote {\bibinfo {title} {Machine learning approaches to evaluate
  heterogeneous treatment effects in randomized controlled trials: a scoping
  review},}\ }\href {\doibase 10.1016/j.jclinepi.2024.111538} {\bibfield
  {journal} {\bibinfo  {journal} {Journal of Clinical Epidemiology}\ }\textbf
  {\bibinfo {volume} {176}} (\bibinfo {year} {2024}),\
  10.1016/j.jclinepi.2024.111538}\BibitemShut {NoStop}%
\bibitem [{\citenamefont {Ma}\ \emph {et~al.}(2014)\citenamefont {Ma},
  \citenamefont {Heritier},\ and\ \citenamefont {Lô}}]{Ma2014}%
  \BibitemOpen
  \bibfield  {author} {\bibinfo {author} {\bibfnamefont {Jun}\ \bibnamefont
  {Ma}}, \bibinfo {author} {\bibfnamefont {Stephane}\ \bibnamefont {Heritier}},
  \ and\ \bibinfo {author} {\bibfnamefont {Serigne~N.}\ \bibnamefont {Lô}},\
  }\bibfield  {title} {\enquote {\bibinfo {title} {On the maximum penalized
  likelihood approach for proportional hazard models with right censored
  survival data},}\ }\href {\doibase
  https://doi.org/10.1016/j.csda.2014.01.005} {\bibfield  {journal} {\bibinfo
  {journal} {Computational Statistics and Data Analysis}\ }\textbf {\bibinfo
  {volume} {74}},\ \bibinfo {pages} {142--156} (\bibinfo {year}
  {2014})}\BibitemShut {NoStop}%
\bibitem [{\citenamefont {Allman}\ \emph {et~al.}(2009)\citenamefont {Allman},
  \citenamefont {Matias},\ and\ \citenamefont {Rhodes}}]{Allman2009}%
  \BibitemOpen
  \bibfield  {author} {\bibinfo {author} {\bibfnamefont {Elizabeth~S}\
  \bibnamefont {Allman}}, \bibinfo {author} {\bibfnamefont {Catherine}\
  \bibnamefont {Matias}}, \ and\ \bibinfo {author} {\bibfnamefont {John~A}\
  \bibnamefont {Rhodes}},\ }\bibfield  {title} {\enquote {\bibinfo {title}
  {Identifiability of parameters in latent structure models with many observed
  variables},}\ }\href@noop {} {\  (\bibinfo {year} {2009})}\BibitemShut
  {NoStop}%
\bibitem [{\citenamefont {Allman}\ \emph {et~al.}(2015)\citenamefont {Allman},
  \citenamefont {Rhodes}, \citenamefont {Stanghellini},\ and\ \citenamefont
  {Valtorta}}]{Allman2015}%
  \BibitemOpen
  \bibfield  {author} {\bibinfo {author} {\bibfnamefont {Elizabeth~S}\
  \bibnamefont {Allman}}, \bibinfo {author} {\bibfnamefont {John~A}\
  \bibnamefont {Rhodes}}, \bibinfo {author} {\bibfnamefont {Elena}\
  \bibnamefont {Stanghellini}}, \ and\ \bibinfo {author} {\bibfnamefont
  {Marco}\ \bibnamefont {Valtorta}},\ }\bibfield  {title} {\enquote {\bibinfo
  {title} {Parameter identifiability of discrete bayesian networks with hidden
  variables},}\ }\href@noop {} {\bibfield  {journal} {\bibinfo  {journal}
  {Journal of Causal Inference}\ }\textbf {\bibinfo {volume} {3}},\ \bibinfo
  {pages} {189--205} (\bibinfo {year} {2015})}\BibitemShut {NoStop}%
\bibitem [{\citenamefont {Gassiat}\ \emph {et~al.}(2016)\citenamefont
  {Gassiat}, \citenamefont {Cleynen},\ and\ \citenamefont
  {Robin}}]{Gassiat2016a}%
  \BibitemOpen
  \bibfield  {author} {\bibinfo {author} {\bibfnamefont {E.}~\bibnamefont
  {Gassiat}}, \bibinfo {author} {\bibfnamefont {A.}~\bibnamefont {Cleynen}}, \
  and\ \bibinfo {author} {\bibfnamefont {S.}~\bibnamefont {Robin}},\ }\bibfield
   {title} {\enquote {\bibinfo {title} {Inference in finite state space non
  parametric hidden markov models and applications},}\ }\href@noop {}
  {\bibfield  {journal} {\bibinfo  {journal} {Statistics and Computing}\
  }\textbf {\bibinfo {volume} {26}},\ \bibinfo {pages} {61--71} (\bibinfo
  {year} {2016})}\BibitemShut {NoStop}%
\bibitem [{\citenamefont {Gassiat}\ and\ \citenamefont
  {Rousseau}(2016)}]{Gassiat2016b}%
  \BibitemOpen
  \bibfield  {author} {\bibinfo {author} {\bibfnamefont {Elisabeth}\
  \bibnamefont {Gassiat}}\ and\ \bibinfo {author} {\bibfnamefont {Judith}\
  \bibnamefont {Rousseau}},\ }\bibfield  {title} {\enquote {\bibinfo {title}
  {Nonparametric finite translation hidden markov models and extensions},}\
  }\href {http://www.jstor.org/stable/43863921} {\bibfield  {journal} {\bibinfo
   {journal} {Bernoulli}\ }\textbf {\bibinfo {volume} {22}},\ \bibinfo {pages}
  {193--212} (\bibinfo {year} {2016})}\BibitemShut {NoStop}%
\bibitem [{\citenamefont {Wieland}\ \emph {et~al.}(2021)\citenamefont
  {Wieland}, \citenamefont {Hauber}, \citenamefont {Rosenblatt}, \citenamefont
  {Tönsing},\ and\ \citenamefont {Timmer}}]{Wieland2021}%
  \BibitemOpen
  \bibfield  {author} {\bibinfo {author} {\bibfnamefont {Franz-Georg}\
  \bibnamefont {Wieland}}, \bibinfo {author} {\bibfnamefont {Adrian~L.}\
  \bibnamefont {Hauber}}, \bibinfo {author} {\bibfnamefont {Marcus}\
  \bibnamefont {Rosenblatt}}, \bibinfo {author} {\bibfnamefont {Christian}\
  \bibnamefont {Tönsing}}, \ and\ \bibinfo {author} {\bibfnamefont {Jens}\
  \bibnamefont {Timmer}},\ }\bibfield  {title} {\enquote {\bibinfo {title} {On
  structural and practical identifiability},}\ }\href {\doibase
  https://doi.org/10.1016/j.coisb.2021.03.005} {\bibfield  {journal} {\bibinfo
  {journal} {Current Opinion in Systems Biology}\ }\textbf {\bibinfo {volume}
  {25}},\ \bibinfo {pages} {60--69} (\bibinfo {year} {2021})}\BibitemShut
  {NoStop}%
\bibitem [{\citenamefont {Watanabe}(2009)}]{Watanabe2009}%
  \BibitemOpen
  \bibfield  {author} {\bibinfo {author} {\bibfnamefont {Sumio}\ \bibnamefont
  {Watanabe}},\ }\href@noop {} {\emph {\bibinfo {title} {Algebraic geometry and
  statistical learning theory}}},\ Vol.~\bibinfo {volume} {25}\ (\bibinfo
  {publisher} {Cambridge university press},\ \bibinfo {year}
  {2009})\BibitemShut {NoStop}%
\bibitem [{\citenamefont {Calderhead}\ and\ \citenamefont
  {Girolami}(2009)}]{Calderhead2009}%
  \BibitemOpen
  \bibfield  {author} {\bibinfo {author} {\bibfnamefont {Ben}\ \bibnamefont
  {Calderhead}}\ and\ \bibinfo {author} {\bibfnamefont {Mark}\ \bibnamefont
  {Girolami}},\ }\bibfield  {title} {\enquote {\bibinfo {title} {Estimating
  bayes factors via thermodynamic integration and population mcmc},}\ }\href
  {\doibase https://doi.org/10.1016/j.csda.2009.07.025} {\bibfield  {journal}
  {\bibinfo  {journal} {Computational Statistics and Data Analysis}\ }\textbf
  {\bibinfo {volume} {53}},\ \bibinfo {pages} {4028--4045} (\bibinfo {year}
  {2009})}\BibitemShut {NoStop}%
\bibitem [{\citenamefont {Watanabe}(2013)}]{Watanabe2013}%
  \BibitemOpen
  \bibfield  {author} {\bibinfo {author} {\bibfnamefont {Sumio}\ \bibnamefont
  {Watanabe}},\ }\bibfield  {title} {\enquote {\bibinfo {title} {A widely
  applicable bayesian information criterion},}\ }\href@noop {} {\bibfield
  {journal} {\bibinfo  {journal} {Journal of Machine Learning Research}\
  }\textbf {\bibinfo {volume} {14}},\ \bibinfo {pages} {867--897} (\bibinfo
  {year} {2013})}\BibitemShut {NoStop}%
\bibitem [{\citenamefont {Drton}\ and\ \citenamefont
  {Plummer}(2017)}]{Drton2017}%
  \BibitemOpen
  \bibfield  {author} {\bibinfo {author} {\bibfnamefont {Mathias}\ \bibnamefont
  {Drton}}\ and\ \bibinfo {author} {\bibfnamefont {Martyn}\ \bibnamefont
  {Plummer}},\ }\bibfield  {title} {\enquote {\bibinfo {title} {A bayesian
  information criterion for singular models},}\ }\href {\doibase
  10.1111/rssb.12187} {\bibfield  {journal} {\bibinfo  {journal} {Journal of
  the Royal Statistical Society Series B: Statistical Methodology}\ }\textbf
  {\bibinfo {volume} {79}},\ \bibinfo {pages} {323--380} (\bibinfo {year}
  {2017})}\BibitemShut {NoStop}%
\bibitem [{\citenamefont {Moral}(2004)}]{Moral2004}%
  \BibitemOpen
  \bibfield  {author} {\bibinfo {author} {\bibfnamefont {Pierre}\ \bibnamefont
  {Moral}},\ }\href@noop {} {\emph {\bibinfo {title} {Feynman-Kac formulae:
  genealogical and interacting particle systems with applications}}}\ (\bibinfo
   {publisher} {Springer},\ \bibinfo {year} {2004})\BibitemShut {NoStop}%
\bibitem [{\citenamefont {Chopin}\ \emph {et~al.}(2020)\citenamefont {Chopin},
  \citenamefont {Papaspiliopoulos} \emph {et~al.}}]{Chopin2020}%
  \BibitemOpen
  \bibfield  {author} {\bibinfo {author} {\bibfnamefont {Nicolas}\ \bibnamefont
  {Chopin}}, \bibinfo {author} {\bibfnamefont {Omiros}\ \bibnamefont
  {Papaspiliopoulos}},  \emph {et~al.},\ }\href@noop {} {\emph {\bibinfo
  {title} {An introduction to sequential Monte Carlo}}},\ Vol.~\bibinfo
  {volume} {4}\ (\bibinfo  {publisher} {Springer},\ \bibinfo {year}
  {2020})\BibitemShut {NoStop}%
\bibitem [{\citenamefont {Janson}(2011)}]{Janson2011}%
  \BibitemOpen
  \bibfield  {author} {\bibinfo {author} {\bibfnamefont {Svante}\ \bibnamefont
  {Janson}},\ }\href {https://arxiv.org/abs/1108.3924} {\enquote {\bibinfo
  {title} {Probability asymptotics: notes on notation},}\ } (\bibinfo {year}
  {2011}),\ \Eprint {http://arxiv.org/abs/1108.3924} {arXiv:1108.3924
  [math.PR]} \BibitemShut {NoStop}%
\bibitem [{\citenamefont {Bach}\ and\ \citenamefont {Jordan}(2003)}]{Bach2003}%
  \BibitemOpen
  \bibfield  {author} {\bibinfo {author} {\bibfnamefont {F.R.}\ \bibnamefont
  {Bach}}\ and\ \bibinfo {author} {\bibfnamefont {M.I.}\ \bibnamefont
  {Jordan}},\ }\bibfield  {title} {\enquote {\bibinfo {title} {Kernel
  independent component analysis},}\ }\href@noop {} {\bibfield  {journal}
  {\bibinfo  {journal} {Journal of Machine Learning Research}\ } (\bibinfo
  {year} {2003})}\BibitemShut {NoStop}%
\bibitem [{\citenamefont {Liu}\ and\ \citenamefont {Nocedal}(1989)}]{Liu1989}%
  \BibitemOpen
  \bibfield  {author} {\bibinfo {author} {\bibfnamefont {Dong~C}\ \bibnamefont
  {Liu}}\ and\ \bibinfo {author} {\bibfnamefont {Jorge}\ \bibnamefont
  {Nocedal}},\ }\bibfield  {title} {\enquote {\bibinfo {title} {On the limited
  memory bfgs method for large scale optimization},}\ }\href@noop {} {\bibfield
   {journal} {\bibinfo  {journal} {Mathematical Programming}\ }\textbf
  {\bibinfo {volume} {45}},\ \bibinfo {pages} {503--528} (\bibinfo {year}
  {1989})}\BibitemShut {NoStop}%
\bibitem [{\citenamefont {Williams}\ and\ \citenamefont
  {Rasmussen}(2006)}]{Rasmussen2006}%
  \BibitemOpen
  \bibfield  {author} {\bibinfo {author} {\bibfnamefont {Christopher~KI}\
  \bibnamefont {Williams}}\ and\ \bibinfo {author} {\bibfnamefont
  {Carl~Edward}\ \bibnamefont {Rasmussen}},\ }\href@noop {} {\emph {\bibinfo
  {title} {Gaussian processes for machine learning}}},\ Vol.~\bibinfo {volume}
  {2}\ (\bibinfo  {publisher} {MIT press Cambridge, MA},\ \bibinfo {year}
  {2006})\BibitemShut {NoStop}%
\bibitem [{\citenamefont {Fukumizu}\ \emph {et~al.}(2007)\citenamefont
  {Fukumizu}, \citenamefont {Bach},\ and\ \citenamefont
  {Gretton}}]{Fukumizu2007}%
  \BibitemOpen
  \bibfield  {author} {\bibinfo {author} {\bibfnamefont {Kenji}\ \bibnamefont
  {Fukumizu}}, \bibinfo {author} {\bibfnamefont {Francis~R.}\ \bibnamefont
  {Bach}}, \ and\ \bibinfo {author} {\bibfnamefont {Arthur}\ \bibnamefont
  {Gretton}},\ }\bibfield  {title} {\enquote {\bibinfo {title} {Statistical
  consistency of kernel canonical correlation analysis},}\ }\href
  {http://jmlr.org/papers/v8/fukumizu07a.html} {\bibfield  {journal} {\bibinfo
  {journal} {Journal of Machine Learning Research}\ }\textbf {\bibinfo {volume}
  {8}},\ \bibinfo {pages} {361--383} (\bibinfo {year} {2007})}\BibitemShut
  {NoStop}%
\bibitem [{\citenamefont {Kanamori}\ \emph {et~al.}(2010)\citenamefont
  {Kanamori}, \citenamefont {Suzuki},\ and\ \citenamefont
  {Sugiyama}}]{Kanamori2010}%
  \BibitemOpen
  \bibfield  {author} {\bibinfo {author} {\bibfnamefont {Takafumi}\
  \bibnamefont {Kanamori}}, \bibinfo {author} {\bibfnamefont {Taiji}\
  \bibnamefont {Suzuki}}, \ and\ \bibinfo {author} {\bibfnamefont {Masashi}\
  \bibnamefont {Sugiyama}},\ }\bibfield  {title} {\enquote {\bibinfo {title}
  {Theoretical analysis of density ratio estimation},}\ }\href@noop {}
  {\bibfield  {journal} {\bibinfo  {journal} {IEICE transactions on
  fundamentals of electronics, communications and computer sciences}\ }\textbf
  {\bibinfo {volume} {93}},\ \bibinfo {pages} {787--798} (\bibinfo {year}
  {2010})}\BibitemShut {NoStop}%
\end{thebibliography}%

\end{document}